\documentclass[a4paper,fleqn]{cas-dc}

\usepackage[numbers]{natbib}
\usepackage{bm}
\usepackage{amsfonts}
\usepackage{amssymb}
\usepackage{amsmath}
\usepackage[noend]{algpseudocode}
\usepackage{algorithmicx,algorithm}
\usepackage{listings}
\DeclareMathOperator*{\argmin}{argmin}
\newtheorem{definition}{Definition}
\newtheorem{assum}{Assumption}
\newtheorem{proposition}{Proposition}

\newtheorem{lemma}{Lemma}
\newtheorem{remark}{Remark}
\newtheorem{theorem}{Theorem}
\newtheorem{corollary}{Corollary}

\newproof{pf}{Proof}
\def\reff#1{{\rm(\ref{#1})}}
\usepackage{subfigure}
\def\tsc#1{\csdef{#1}{\textsc{\lowercase{#1}}\xspace}}
\tsc{WGM}
\tsc{QE}
\tsc{EP}
\tsc{PMS}
\tsc{BEC}
\tsc{DE}
\begin{document}
\let\WriteBookmarks\relax
\def\floatpagepagefraction{1}
\def\textpagefraction{.001}
\shorttitle{Projection Neural Network for a Class of Sparse Regression Problems with Cardinality Penalty}
\shortauthors{}

\title [mode = title]{Projection Neural Network for a Class of Sparse Regression Problems with Cardinality Penalty}
\tnotemark[1]                   

\tnotetext[1]{This work is supported in part by the National Natural Science Foundation of China under Grant 11871178 and Grant 61773136.}

\author[1]{Wenjing Li}[orcid=0000-0002-4351-6360]
\ead{liwenjingsx@163.com}


\address[1]{School of Mathematics, Harbin Institute of Technology, Harbin 150001,  China}

\author[1,2]{Wei Bian}[orcid=0000-0003-4252-047X]
\cormark[1]
\ead{bianweilvse520@163.com}


\address[2]{Institute of Advanced Study in Mathematics, Harbin Institute of Technology, Harbin 150001,  China}


\cortext[cor1]{Corresponding author}


\begin{abstract}
In this paper, we consider a class of sparse regression problems, whose objective function is the summation of a convex loss function and a cardinality penalty. By constructing a smoothing function for the cardinality function, we propose a projection neural network and design a correction method for solving this problem. The solution of the proposed neural network is unique, global existent, bounded and globally Lipschitz continuous. Besides, we prove that all accumulation points of the proposed neural network have a common support set and a unified lower bound for the nonzero entries. Combining the proposed neural network with the correction method, any corrected accumulation point is a local minimizer of the considered sparse regression problem. Moreover, we analyze the equivalence on the local minimizers between the considered sparse regression problem and another sparse regression problem. Finally, some numerical experiments are provided to show the efficiency of the proposed neural network in solving some sparse regression problems.
\end{abstract}




\maketitle

\section{Introduction}
Sparse regression problem is a core problem in many engineering and scientific fields, such as compressed sensing \cite{Candes2006}, high-dimensional statistical learning \cite{B2014High}, variable selection \cite{Liu2007Variable}, imaging decomposition \cite{Soubies2015A}. Sparse regression problem is to find the sparsest solution of a linear or nonlinear regression model. One typical example of sparse regression problem is to find the sparsest solution of the following linear system
\begin{equation*}\label{linear model}
b=Ax+\varepsilon,
\end{equation*}
where $A\in\mathbb{R}^{m\times n}$ is the design matrix, $b\in\mathbb{R}^m$ is the response vector, $\varepsilon\in\mathbb{R}^m$ is the noise vector. Best subset selection and Lasso are popular methods for selection and estimation of the parameters in a linear regression model under squared-error loss \cite{Hastie2017}. Cardinality function on $\mathbb{R}^n$ is also called the $L_0$-norm and denoted by $\|\cdot\|_0$. For $x\in\mathbb{R}^n$, \begin{equation*}
\|x\|_0:=|\{i=1,2,\ldots,n:x_i\neq0\}|,
\end{equation*} where $|S|$ is the cardinality of set $S$. $x\in\mathbb{R}^n$ is called sparse if $\|x\|_0\ll n$. Cardinality function is an effective concept for controlling the sparsity of data and plays an important role in sparse regression problems \citep{H2015DC}, since it penalizes the number of nonzero elements directly and can increase the accurate identification rate of the estimator on the important predictors\citep{Nikolova2016Relationship}. However, it is known that the sparse regression problems with cardinality penalty are NP-hard in general \citep{Natarajan1995Sparse}. So, the development of algorithms for solving this kind of sparse regression problem is still a challenge up to now.

When the prior about the noise or sparsity of the solution is unknown, the sparse regression problem with cardinality penalty is usually formulated as the following form \cite{Soubies2015A}
\begin{equation*}
\begin{split}
&\min_{x\in\mathbb{R}^n}\quad\frac{1}{2}\|Ax-b\|^2+\lambda\|x\|_0,\\
\end{split}
\end{equation*}
where $\lambda>0$ is a hyperparameter characterizing the trade-off between data fidelity and sparsity.

For solving a wider class of sparse regression problems, we consider the following sparse regression problem with cardinality penalty
\begin{equation}\label{optimization model}
\begin{split}
&\min\quad f(x)+\lambda\|x\|_0\\
&~\text{s.t.}~\quad x\in\mathcal{X}:=\{x\in\mathbb{R}^{n}:\bm{0}\leq{x}\leq\upsilon\},
\end{split}
\end{equation}
where $\upsilon\in\mathbb{R}_{+}^{n}$, $\lambda>0$, $f:\mathbb{R}^n\rightarrow\mathbb{R}$ is a continuously differentiable convex function and $\nabla{f}$ is locally Lipschitz continuous. In (\ref{optimization model}), $f$ is the loss function to guarantee the match of the data fitting and $\lambda\|x\|_0$ is the penalty to promote the sparsity of the solution. The considered problem \reff{optimization model} is a class of nonconvex and discontinuous optimization problems.

Different from the existing methods for solving the relaxation problems of the cardinality penalty problems, such as $L_p(0<p<1)$-penalized and $L_1$-penalized sparse regression problems \citep{Chen2014Complexity,Bian2015Complexity,Liu2016}, this paper focuses on the original cardinality penalty problem. Directly solving the regression problems with cardinality penalty is an interesting topic. At first, Mohimani, Babaie-Zadeh and Juttena proposed a fast algorithm for overcomplete sparse decomposition based on smoothed $L_0$ norm \citep{Jutten2008A}. Further, for $L_0$-penalized least-square problems, Jiao, Jin and Lu developed a primal dual active set with continuation (PDASC) algorithm \citep{Jiao2014}. Based on DC programming and algorithms, Le Thia et al. offered a unifying nonconvex approximation approach to solve sparse regression problems with finite DC loss function and cardinality penalty \citep{H2015DC}. Recently, a smoothing proximal gradient algorithm is proposed for nonsmooth convex regression with cardinality penalty \citep{Bian2019Sparse}. Moreover, there are also some recent research progress in \citep{Pan2017PAMI,Cai2019,Xiong2019}. The above algorithms for solving the regression problems with cardinality penalty are all iterative algorithms and to the best of our knowledge, there is no neural network based on circuit implementation to solve such problems so far.

In scientific and engineering fields, real-time solving is necessary for some optimization problems, so neural networks have been studied gradually. Dynamic algorithms and iterative algorithms are two important kinds of algorithms for solving optimization problems and they can promote each other \cite{Osher2016,Attouch2018MP}. Dynamic algorithms have some particular advantages. For instance, it is not necessary to select the search direction or step size in implementation, and some differential equations have a promoting effect on the acceleration of iterative algorithms \citep{Su2016,Attouch2016,Attouch2020}. Neural networks modeled by a class of dynamical algorithms not only own the advantages of dynamic methods mentioned above, but also own their particular superiorities. Neural network can be implemented physically by hardware such as integrated circuits and hence it can solve optimization problems faster in running time at the order of magnitude \cite{Xia2008A,Gao2009A}. Some classical neural networks were designed to solve the linear and nonlinear programming in \citet{Hopfield1985}, \citet{Tank1988Simple} and \citet{Kennedy1988Neural}. Invoked by these work, many researchers developed different neural networks for solving various optimization problems. There are many interesting results on hardware implementation of neural networks \citep{Clemente2016,Chen2020} and solving the continuous convex and nonconvex optimization problems by neural networks \citep{Bian2013Neural,Yan2017A,Le2017A,BianNN2018}. Sparse regression problem \reff{optimization model} is a class of discontinuous and nonconvex optimization problems and the existing neural networks cannot be directly used to solve such problem. As a result, it is necessary to design neural networks modeled by differential equations to solve problem \reff{optimization model}, which can also further extend the study of neural network for solving the discontinuous and nonconvex optimization problems.

In order to overcome the discontinuity of cardinality pe- nalty, some researchers have designed some continuous nonconvex penalties to relax it, such as the truncated $L_1$ penalty \citep{Shen2012Likelihood}, hard thresholding penalty \citep{Zheng2014High}, bridge $L_p(0<p<1)$ penalty \citep{Foucart2009Sparsest}, capped-$L_1$  penalty \citep{Zhang2013Multi}, smoothly clipped absolute deviation (SCAD) penalty \citep{Fan2001Variable}, minimax concave penalty (MCP) \citep{Zhang2010NEARLY}, continuous exact $L_0$ penalty (CEL0) \citep{Soubies2015A}, etc. Among them, $L_{1/2}$ quasi-norm is an important regularization term in the study of compressive sensing \citep{Xu2012L1}. However, for the theoretical analysis of projection neural networks, most of the above penalties are not applicable because of non-differentiability, non-Lipschitz continuity, parameter selection and approximation properties. For the detailed, we will give it in Remark 2. In recent years, based on smoothing techniques, neural networks can gradually be used to solve non-Lipschitz and nonconvex optimization problems \citep{Wei2012Smoothing,Bian2014Neural,Li2019}. Inspired by smoothing techniques, according to special geometric properties of cardinality penalty and the smoothness requirements of projection neural networks for theoretical analysis, we design a smoothing function of cardinality penalty in this paper, which is also a novel nonconvex relaxation of cardinality penalty.

Different from the penalty method for the constrained optimization problems, projection neural network can reduce the complexity of circuit components for solving problem \reff{optimization model}. Then, by designing a smoothing function of cardinality penalty, we will propose a projection neural network and a correction method for solving sparse regression problem \reff{optimization model}.
The main contributions of this paper are as follows.
\begin{itemize}
\item The considered problem (\ref{optimization model}) is a class of sparse regression problems with the desirable sparsity penalty $\|\cdot\|_0$, which is a more general form of \citep{Soubies2015A,Soubies2017A} and different from its some relaxation problems in \citep{Chen2014Complexity,Bian2015Complexity,Liu2016}. Under squared-error loss, the equivalence of these relaxation problems to (\ref{optimization model}) needs some conditions \cite{Candes2006,Foucart2009Sparsest,2011Equivalence}. And the relationships are not clear when $f$ is a general convex function. In this paper, we consider the problem (\ref{optimization model}) directly. Though we solve it by a smoothing approximation model of it, we prove the optimality properties of its accumulation points to (\ref{optimization model}).
\item A smoothing function of cardinality function is designed, which is a nonconvex relaxation of cardinality function but more appropriate for the analysis of neural network than the existing relaxation penalties in \citep{Soubies2015A,Shen2012Likelihood,Zheng2014High,Foucart2009Sparsest,Zhang2013Multi,Fan2001Variable,Zhang2010NEARLY}.
\item A projection neural network with smoothing technique is proposed for solving the discontinuous and nonconvex optimization problem (\ref{optimization model}), which is the first differential equation system for solving (\ref{optimization model}) to the best of our knowledge, and whose analysis is different from the existing neural networks for solving the continuous optimization problems. 
\item In order to expand the applicable range of the proposed neural network, we further consider the following sparse regression model
\begin{equation}\label{two side0}
\begin{split}
&\min\quad f(x)+\lambda\|x\|_0\\
&~\text{s.t.}~\quad x\in{\mathcal{Y}}:=\{x\in\mathbb{R}^{n}:-l\leq{x}\leq u\},
\end{split}
\end{equation}
where $u,l\in\mathbb{R}^n_+$, $\lambda>0$ and $f$ is defined as in \reff{optimization model}. Using variable splitting $x=x_+-x_-$, we prove that in the sense of local minimizers, problem \reff{two side0} can be equivalently converted to the following special case of sparse regression problem \reff{optimization model}
\begin{equation}\label{oneside0}
\begin{aligned}
&\min &&f(x_+-x_-)+\lambda\|x_+\|_0+\lambda\|x_-\|_0\\
&~\text{s.t.}~ &&x_+\in {\mathcal{X}_1}:=\{x\in\mathbb{R}^{n}:{\bm 0}\leq{x}\leq u\},\\
&\,&&x_-\in{\mathcal{X}_2}:=\{x\in\mathbb{R}^{n}:\bm{0}\leq{x}\leq l\}.
\end{aligned}
\end{equation}
\item Through numerical experiments, we have confirmed that the proposed neural network is insensitive to initializations and has better performance than the methods in \citep{Hastie2017,Bian2015Complexity,Zhao2020,Feng2017} for some specific experiments.
\end{itemize}

We organize the remaining part of this paper as follows. In Section \ref{section2}, some preliminary results are given. In Section \ref{section3}, we define a smoothing function for the cardinality penalty in \reff{optimization model} and analyze its some necessary properties. In Section \ref{section4}, we propose a projection neural network and analyze the properties of its solution in solving problem \reff{optimization model}. Moreover, a method to correct the accumulation points of the proposed neural network is designed in order to obtain a local minimizer of \reff{optimization model} with better sparsity. In Section \ref{section6}, the equivalence between the local minimizers of \reff{two side0} and \reff{oneside0} are proved. Finally, some numerical examples are illustrated in Section \ref{section7} to show the good performance of the proposed network in solving problems \reff{optimization model} and \reff{two side0}. 

\textbf{Notations:} $\mathbb{R}_{+}^n$ denotes the set composed by all $n$ dimensional nonnegative vectors. For $x\in\mathbb{R}^n$, $\|x\|:=\|x\|_2=\left(\sum_{i=1}^{n}x_i^2\right)^\frac{1}{2}$ and $\|x\|_\infty=\max\{|x_i|:i=1,2,\ldots,n\}$. Denote $\textbf{k}_n=(k,k,\ldots,k)\in\mathbb{R}^n$. For $x\in\mathbb{R}^n$ and an index set $\Pi\subseteq\{1,2,\ldots,n\}$,
$|\Pi|$ denotes the number of elements in $\Pi$,
$x_{\Pi}:=(x_{i_1},x_{i_2},\ldots,x_{i_{|\Pi|}})$ with $i_1,i_2,\ldots,i_{|\Pi|}\in\Pi$ and $i_1<i_2<\ldots<i_{|\Pi|}$. For an $x\in\mathbb{R}^n$ and $\delta>0$, ${B}_{\delta}(x)$ denotes the open ball in $\mathbb{R}^n$ centered at $x$ with radius $\delta$.

\section{Preliminaries}\label{section2}
\indent In this section, we will introduce some necessary definitions and properties that will be used in what follows.

\begin{proposition}\label{limit}\citep{Coddington1984Theory}
Let $g:[0,+\infty)\rightarrow\mathbb{R}$ be a continuous function. If $\int_0^{+\infty}|g(s)|ds< +\infty$ and $g$ is uniformly continuous on $[0,+\infty)$\footnote{We call $g:[0,+\infty)\rightarrow\mathbb{R}$ uniformly continuous on $[0,+\infty)$, if for every $\epsilon>0$, there exists $\delta>0$ such that for every $s,t\in[0,+\infty)$ with $|s-t|<\delta$, we have that $|g(s)-g(t)|<\epsilon$.}, then $\lim_{s\rightarrow+\infty}g(s)=0$.
\end{proposition}

For a nonempty, closed and convex set $\Omega\subseteq{\mathbb{R}}^{n}$, the projection operator to $\Omega$ at $x$ is defined by
\begin{equation*}
P_{\Omega}(x)=\argmin_{u\in{\Omega}}\|u-x\|^2,
\end{equation*}
which owns the following properties.
\begin{proposition}\label{projection operator property}\citep{Kinderlehrer1980An}
Assume $\Omega$ is a closed and convex subset of ${\mathbb{R}}^{n}$. Then, $P_{\Omega}(\cdot)$ owns the following properties:
\begin{align*}
\begin{split}
\left\langle{w-P_{\Omega}(w), P_{\Omega}(w)-u}\right\rangle\geq0,\quad~~\forall w \in {\mathbb{R}}^{n}, u \in \Omega;\\
\|P_{\Omega}(u)-P_{\Omega}(w)\|\leq\|u-w\|,\quad~~\forall u, w \in {\mathbb{R}}^{n}.
\end{split}
\end{align*}
\end{proposition}

\begin{proposition}\label{minimizer}\citep{Kinderlehrer1980An}
If $f:\Omega\rightarrow\mathbb{R}$ is a convex and differentiable function on the closed and convex set $\Omega$, then \begin{equation*}
x^*=P_{\Omega}\left(x^*-\nabla f(x^*)\right)
\end{equation*} if and only if $x^*$ is a global minimizer of $f$ in ${\Omega}$.
\end{proposition}

Definitions of globally and locally Lipschitz continuous functions can be found in \citet{Clarke1983Optimization}.

Let $D$ be an open set in $\mathbb{R}^{n+1}$ with an element of $D$ written as $(x,t)$ and $G:D\rightarrow\mathbb{R}^n$ be a continuous function. For a nonautonomous real-time system modeled by a differential equation
\begin{align*}
\dot{x}(t)=G(x(t),t),
\end{align*} we call $x:[0,T)\rightarrow\mathbb{R}^{n}$ one of its solutions, if $x$ is continuously differentiable on $(0,T)$, $(x(t),t)\in D$ for $t\in [0,T)$ and $x$ satisfies it on $(0,T)$ \citep{Hale1980}.

Since $\|\cdot\|_0$ in \reff{optimization model} is discontinuous, we introduce the smoothing method into neural network and use the smoothing function defined as follows, which is restricted to a closed and convex subset of $\mathbb{R}^n$.
\begin{definition}\label{smoothing function definition}
Let $h: \mathbb{R}^n\rightarrow\mathbb{R}$ be a function and $\Omega$ be a closed and convex subset of ${\mathbb{R}}^{n}$. We call $\tilde h: \mathbb{R}^n\times(0,+\infty)\rightarrow\mathbb{R}$ a smoothing function of $h$ on $\Omega$, if $\tilde{h}(\cdot, \mu)$ is differentiable on $\mathbb{R}^n$ for any fixed $\mu>0$ and $\lim_{\mu\downarrow{0}}\tilde{h}(x,\mu) = h(x)$ holds for any $x\in\Omega$.
\end{definition}
\section{Smoothing function of $\|x\|_0$ on $\mathcal{X}$}\label{section3}
In this section, we design a smoothing function of $\|x\|_0$ on $\mathbb{R}^n_+$ and give its some necessary properties, which will be used in the analysis on the proposed neural network for solving \reff{optimization model}.

\begin{figure}
\centering
\subfigure[]{\includegraphics[width=1.585in]{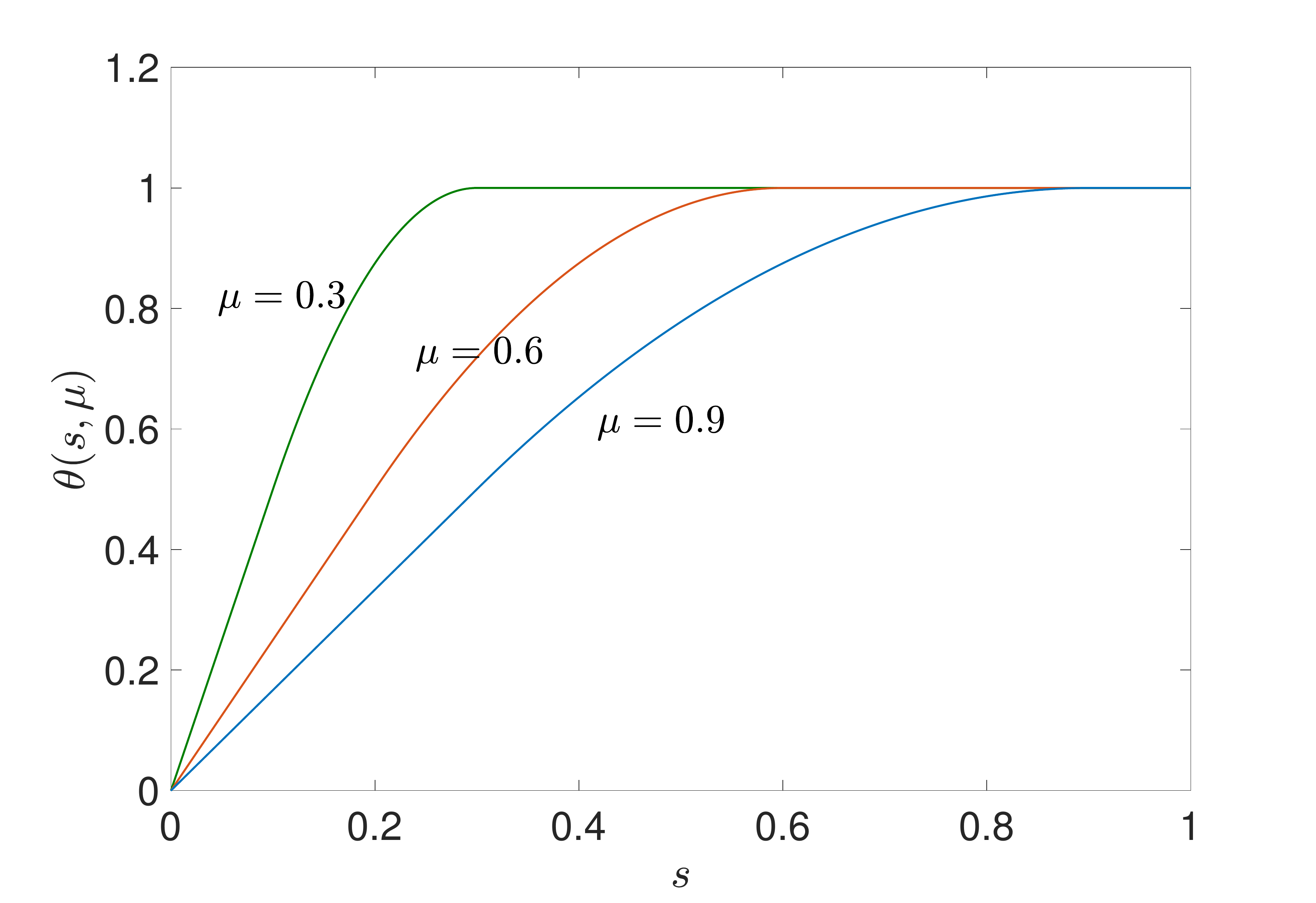}
\label{fig:smoothing}}
\hfil
\subfigure[]{\includegraphics[width=1.585in]{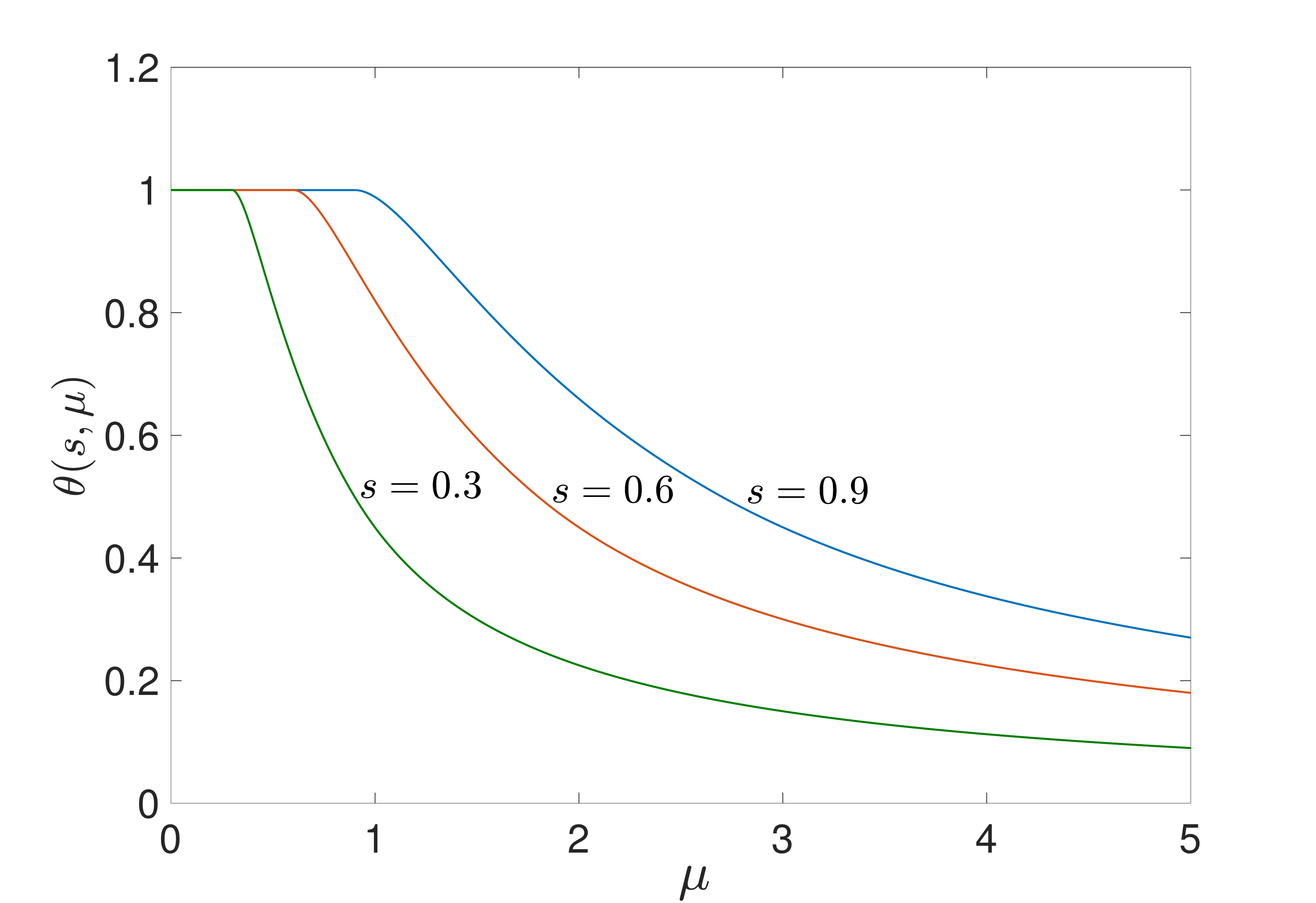}
\label{fig:smoothing2}}
\caption{(a) $\theta(\cdot,\mu)$ with $\mu=0.3,0.6,0.9$; (b) $\theta(s,\cdot)$ with $s=0.3,0.6,0.9$.} \label{fig:smoothing function}
\end{figure}

Define
\begin{equation}\label{sftheta}
\Theta(x,\mu)=\sum_{i=1}^{n}\theta(x_i,\mu),
\end{equation}
where
\begin{gather}\label{smoothing}
\theta(s, \mu)=\left\{
\begin{split}
&\frac{3}{2\mu}s~~~~~~~~~~\qquad \qquad {\rm if}~{s}<\frac{1}{3}\mu,\\
&-\frac{9}{8\mu^2}(s-\mu)^2+1~~ {\rm if}~\frac{1}{3}\mu\leq{s}\leq\mu,\\
&1~~~~~~~~~~~~~~~~~~~~\qquad\quad {\rm if}~{s}>\mu.
\end{split}
\right.
\end{gather}
For some fixed $\mu>0$, the presentation of $\theta(\cdot,\mu)$ on $[0,1]$ is pictured in Fig. \ref{fig:smoothing function} (a). Meantime, the presentation of $\theta(s,\cdot)$ on $(0,5]$ for some fixed $s>0$ is shew in Fig. \ref{fig:smoothing function} (b).

For any $x\in\mathbb{R}^n$ and $\mu>0$, the gradient of $\Theta(x, \mu)$ is denoted by $\nabla\Theta(x, \mu)$ and \begin{equation*}
\nabla\Theta(x, \mu)=\left(\begin{split}
&\nabla_x\Theta(x, \mu)\\
&\nabla_\mu\Theta(x, \mu)
\end{split}\right).
\end{equation*} Here, $\nabla_x\Theta(x, \mu)$ is the gradient of $\Theta(x, \mu)$ at $x$ with a fixed $\mu$ and $\nabla_\mu\Theta(x, \mu)$ is the gradient of $\Theta(x, \mu)$ at $\mu$ with a fixed $x$.

\begin{definition}\label{stationary}
For a fixed $\mu>0$, we call $x^s_\mu$ a $\mu$-stationary point of the following smoothing optimization problem
\begin{equation}\label{smoothing optimization model}
\begin{split}
&\min\quad f(x)+\lambda\Theta\left(x,\mu\right)\\
&~\text{s.t.}~\quad x\in\mathcal{X},
\end{split}
\end{equation}
where $f$ and $\mathcal{X}$ are the same as them in \reff{optimization model},
if $x^s_\mu$ satisfies
\begin{align*}
x^s_\mu=P_{\mathcal{X}}\big[x^s_\mu-\nabla{f}(x^s_\mu)-\lambda\nabla_x\Theta(x^s_\mu,\mu )\big].
\end{align*}
\end{definition}

Let $\bar{x}_\mu^s$ be a global minimizer of \reff{smoothing optimization model}, then when $\mu\downarrow0$, any accumulation point of $\{\bar{x}_\mu^s:\mu>0\}$ is a global minimizer of \reff{optimization model}. Based on the above construction and following analysis of $\Theta(x, \mu)$, we will show that when $\mu$ is small enough, if $\left\{j\in\{1,2,\ldots,n\}:{x^s_\mu}_j\in\left[\frac{1}{6}\mu,\frac{1}{2}\mu\right)\right\}=\emptyset$, then $x^s_\mu$ is a local minimizer of \reff{optimization model} in the following Remark \ref{sta}.
\begin{proposition}\label{prop1}
$\Theta(x,\mu)$ is a smoothing function of $\|x\|_0$ on $\mathbb{R}^n_+$ defined in Definition \ref{smoothing function definition} and satisfies the following properties.
\begin{enumerate}[(i)]
\item For any fixed $\mu>0$, $\Theta(\cdot,\mu)$ is continuously differentiable and $\nabla_x\Theta(\cdot, \mu)$ is globally Lipschitz continuous on $\mathbb{R}^n$.
\item For any fixed $x\in\mathbb{R}^n$, $\Theta(x,\cdot)$ is continuously differentiable and $\nabla_{\mu}\Theta(x,\cdot)$ is locally Lipschitz continuous on $(0,+\infty)$.
\item For $\bar{\mu}\geq\underline{\mu}>0$, $\nabla_{x}\Theta$ and $\nabla_{\mu}\Theta$ are bounded and globally Lipschitz continuous on $\mathcal{X}\times[\underline{\mu},\bar{\mu}]$.
\end{enumerate}
\end{proposition}
\begin{pf}
See Appendix \ref{a1}.
\end{pf}

\section{Neural network}\label{section4}
In this section, we will propose a projection neural network and a correction method when it is needed. Some dynamic and optimal properties of the proposed neural network and method for solving \reff{optimization model} are also analyzed. 

Since $\mathcal{X}$ is bounded and $\nabla{f}$ is locally Lipschitz continuous in problem \reff{optimization model}, it is naturally satisfied that $\exists~L_f>0$, $s.t.~\sup_{x\in{\mathcal{X}}}\|\nabla{f(x)}\|_\infty\leq L_f$ and $\nabla{f}$ is globally Lipshcitz continuous on $\mathcal{X}$. Throughout this paper, we need the value of $L_f$ to support the theoretical results of this paper and the following proposed neural network is qualified for the situation where $L_f$ is available. Moreover, we need the following parameters.
Denote $\bar{v}=\|\upsilon\|_\infty$ and
$\underline{v}=\min\{\upsilon_i: \upsilon_i\neq0,i=1,2,\ldots,n\}$.

Based on the smoothing function $\Theta$ of $\|x\|_0$ on $\mathbb{R}_+^n$ designed in Section \ref{section3}, we propose a projection neural network (Algorithm 1) modeled by a differential equation to solve \reff{optimization model}.
\begin{algorithm}
\caption{Projection neural network for problem (\ref{optimization model})}
{\bf Initialization:} $x_0\in\mathcal{X}$.\\	
\begin{small}
\begin{equation}\label{Neural Network}
\left\{
\begin{split}
&\dot{x}(t)=\gamma\Bigg[-x(t)+P_{\mathcal{X}}\Big[x(t)-\nabla{f}\left(x(t)\right)-\lambda\nabla_x\Theta\left(x(t),\mu(t)\right)\Big]\Bigg]\\
&x(0)=x_0,
\end{split}
\right.
\end{equation}
\end{small}where $\gamma$ is a given positive parameter, $\mu(t)=\frac{1}{2}(\frac{\alpha_0}{(t+1)^\beta}+\mu^*)$ with any given positive parameters $\alpha_0$ and $\beta$ and a positive parameter $\mu^*$ satisfying Assumption \ref{choice}.
\end{algorithm}
\begin{assum}\label{choice}
\begin{equation*}
0<{\mu^*}<\min\left\{\underline{v},\frac{3\lambda}{2(\bar{v}+L_f)},\frac{2\lambda}{{n}L_f}\right\}.
\end{equation*}
\end{assum}

According to the expression of the nonautonomous term $\mu(t)$ in \reff{Neural Network}, it is the solution of the following autonomous differential equation
\begin{equation}\label{mugradientexpression}
\left\{
\begin{split}
&\dot{\mu}(t)=-\beta{\alpha_0}^{-1/\beta}(2\mu(t)-\mu^*)^{1+\frac{1}{\beta}}/2\\
&\mu(0)=\left(\alpha_0+\mu^*\right)/2.
\end{split}
\right.
\end{equation}Combining equations (\ref{Neural Network}) and (\ref{mugradientexpression}), neural network (\ref{Neural Network}) can also be seen as one of traditional neural networks modeled by differential equations with $(x,\mu)$ as variables.

\begin{figure}
	\begin{center}
		\includegraphics[width=0.5\textwidth]{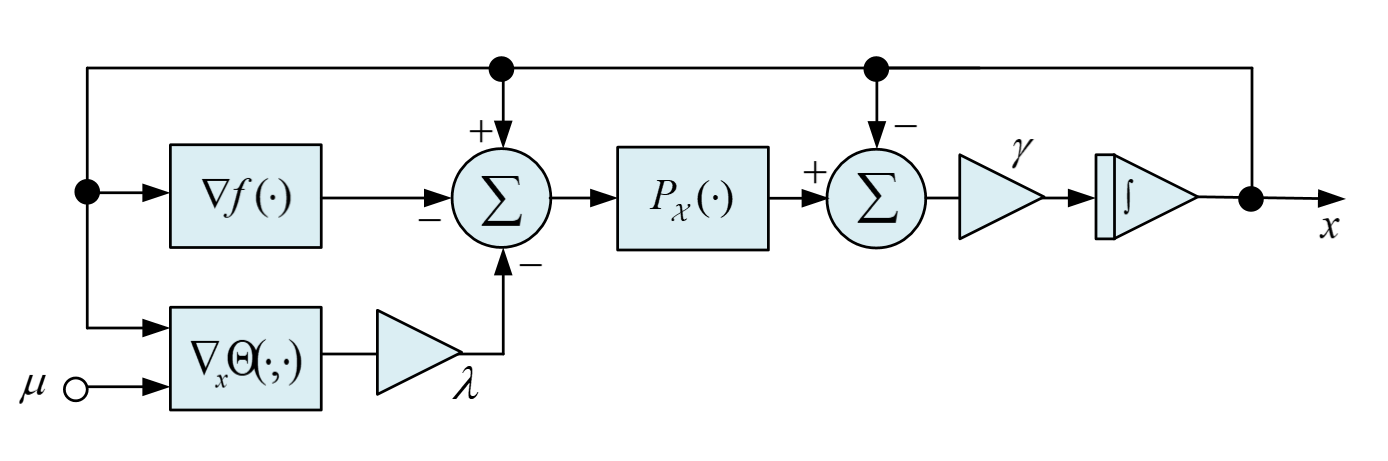}
	\end{center}
	\caption{Schematic block structure of neural network \reff{Neural Network}.}
	\label{fig:second}
\end{figure}

Similar to the explanation in \citet{Bian2013Neural}, neural network \reff{Neural Network} can be implemented by the schematic block structure in Fig. \ref{fig:second}.
\begin{remark}
In (\ref{Neural Network}), $\mu(t)$ can be reformulated by $\mu(t)=\frac{1}{2}(\alpha(t)+\mu^*)$, where $\mu^*$ satisfies Assumption \ref{choice} and $\alpha: [0,+\infty)\rightarrow[0,+\infty)$ is a differentiable decreasing function satisfying 
\begin{itemize}
\item $\lim_{t\rightarrow+\infty}\alpha(t)=0$;
\item $\alpha'(t)$ is bounded and globally Lipschitz continuous on $[0,+\infty)$.
\end{itemize} For example, we can also choose $\alpha(t)=\alpha_0e^{-\beta t}$, where $\alpha_0$ and $\beta$ are any given positive parameters. The convergence and validity of \reff{Neural Network} are not affected by the different selection of $\gamma$, $\alpha_0$, $\beta$ and $\mu^*$.
\end{remark}
\begin{remark}
We compare the relaxation functions of the cardinality function with $\theta$ as follows. In order to analyze the global existence and uniqueness of differential equations, $\theta$ is constructed to be continuously differentiable and its gradient is Lipschitz continuous. But the truncated $L_1$ \cite{Shen2012Likelihood} and capped-$L_1$ \cite{Zhang2013Multi} are non-differentiable and the gradient of bridge $L_p(0<p<1)$ \cite{Foucart2009Sparsest} is non-Lipschitz continuous. $\theta$ is a one-parameter function and approaches to the cardinality function on $\mathbb{R}_+$ as the parameter tends to $0$. But SCAD \cite{Fan2001Variable} and MCP \cite{Zhang2010NEARLY} have two parameters. Moreover, hard thresholding \cite{Zheng2014High} and SCAD can not approach to cardinality function with proper adjustment on the parameters. CEL0 \cite{Soubies2015A} is the MCP with a specific choice of parameters and can be seen as a one-parameter function. $\theta$ is a linear function near $0$, but CEL0 is a quadratic function near $0$, which is likely to result in a heavier computation in the implementation of the neural network and a smaller lower bound for the nonzero elements of the local minimizers than $\theta$ used in this paper. Therefore, $\theta$ is more appropriate for neural network model than these relaxation functions.
\end{remark}
\subsection{Basic dynamic properties of \reff{Neural Network}}\label{section3.1}
In this subsection, we will analyze some basic properties of the solution of neural network \reff{Neural Network}, including its global existence and uniqueness given in Theorem \ref{theorem1} and some basic convergence properties shown in Lemma \ref{lemma1}.

For simplicity of notation, we use $x_{t}$ and $\mu_t$ to denote $x(t)$ and $\mu(t)$ respectively throughout this paper.

\begin{theorem}\label{theorem1}
For any initial point $x_{0}\in{\mathcal{X}}$, there exists a unique global solution $x_t\in C^{1,1}\left([0,+\infty);\mathbb{R}^n\right)$\footnote{$C^{1,1}\left([0,+\infty);\mathbb{R}^n\right)$ denotes the set of all continuously differentiable and globally Lipschitz continuous functions defined on $[0,+\infty)$ and valued in $\mathbb{R}^n$.} to neural network \reff{Neural Network}. Moreover, $x_t\in{\mathcal{X}}$ for any ${t}\in[0,+\infty)$ and $\dot{x}_t$ is globally Lipschitz continuous on $[0,+\infty)$.
\end{theorem}
\begin{pf}
Since the right-hand function of neural network \reff{Neural Network} is continuous with respect to $x$ and $t$, there exists at least one solution $x_t$ to \reff{Neural Network} \citep[pp.14, Theorem 1.1]{Hale1980}.
	
Assume $x_t$ is a solution of \reff{Neural Network} which is not global and its maximal existence interval is $[0,T)$ with $T>0$.	
	
First of all, we prove that $x_{t}\in{\mathcal{X}}, \forall{t}\in[0, T)$. Rewrite \reff{Neural Network} as $\dot{x}_{t}+\gamma{x}_{t}=\gamma h(t)$, where \begin{equation*}
h(t)=P_{\mathcal{X}}\left[x_t-\nabla{f}\left(x_t\right)-\lambda\nabla_x\Theta\left(x_t,\mu_t\right)\right]
\end{equation*} is a continuous function. Since $(e^{\gamma t}x_{t})'=\gamma e^{\gamma t}h(t)$, we have \begin{equation*}
e^{\gamma t}x_{t}-x_{0}=\int_{0}^{t}\gamma e^{\gamma s}h(s)ds,
\end{equation*}which means
\begin{equation}\label{convex1}
x_{t}=e^{-\gamma t}x_{0}+(1-e^{-\gamma t})\int_{0}^{t}\frac{\gamma e^{\gamma s}}{e^{\gamma t}-1}h(s)ds.
\end{equation}
Since $h(s)$ and $\frac{\gamma e^{\gamma s}}{e^{\gamma t}-1}$ are continuous on $[0,t)$, $\frac{\gamma e^{\gamma s}}{e^{\gamma t}-1}>0$, $\int_{0}^{t}\frac{\gamma e^{\gamma s}}{e^{\gamma t}-1}ds=1$ and $h(s)\in{\mathcal{X}}$, $\forall{s}\in[0,t]$, we have for any $t\in[0, T)$,
\begin{align}\label{convex2}
\int_{0}^{t}\frac{\gamma e^{\gamma s}}{e^{\gamma t}-1}h(s)ds\in{\mathcal{X}}.
\end{align}
Combining \reff{convex1} with \reff{convex2}, by the convexity of $\mathcal{X}$, we deduce for any $t\in[0,T)$, $x_{t}\in\mathcal{X}$.
Hence, in view of the compactness of $\mathcal{X}$, we obtain that $x_t$ and $\dot{x}_t$ are bounded on $[0,T)$.
By \citet[pp.16, Lemma 2.1]{Hale1980}, $x_t$ can be extended, which leads to a contradiction. As a consequence, $x_t$ is global existent and $x_t\in\mathcal{X}$ for any $t\in[0,+\infty)$. Owning to the boundedness of $\mathcal{X}$, by the structure of $\dot{x}_t$ in \reff{Neural Network}, we obtain that $\dot{x}_t$ is bounded on $[0,+\infty)$, which implies $x_t\in C^{1,1}\left([0,+\infty);\mathbb{R}^n\right)$.
	
Next, we prove the uniqueness of the solution of \reff{Neural Network}. Let $x$ and $\hat{x}$ be two solutions of neural network \reff{Neural Network} with initial point $x_{0}\in\mathcal{X}$ and we suppose there exists a $\hat{t}>0$ such that $x_{\hat{t}}\neq\hat{x}_{\hat{t}}$. From the continuity of $x_t$ and $\hat{x}_t$, there exists a $\delta>0$ such that $x_{t}\neq\hat{x}_{t}, \forall{t\in[\hat{t},\hat{t}+\delta]}$. Define $$\xi(x,\mu)=\gamma\Bigg[-x+P_{\mathcal{X}}\Big[x-\nabla f(x)-\lambda\nabla_x\Theta\left(x,\mu\right)\Big]\Bigg].$$
Since $\nabla f(\cdot)$ and $\nabla_x\Theta(\cdot,\mu)$ are globally Lipschitz continuous on $\mathcal{X}$, by the global Lipshcitz continuity of $P_\mathcal{X}(\cdot)$ given in Proposition \ref{projection operator property}, we have that $\xi(\cdot,\mu)$ is globally Lipschitz continuous on $\mathcal{X}$ for any fixed $\mu>0$.
Then, from the continuity of $x_t$, $\hat{x}_t$ and $\mu_{t}$ on $[0,\hat{t}+\delta]$, it follows that there exists an $L_\xi>0$ such that
\begin{align*}
\|\xi(x_t,\mu_t)-\xi(\hat{x}_t,\mu_t)\|\leq{L_\xi}\|x_t-\hat{x}_t\|,\quad \forall t\in[0,\hat{t}+\delta].
\end{align*}
Thus, for any $t\in[0,\hat{t}+\delta]$, we have
\begin{align}\label{integrate}
\frac{d}{dt}\|x_{t}-\hat{x}_{t}\|^{2}\leq2{L_\xi}\|x_{t}-\hat{x}_{t}\|^{2}.
\end{align}
Integrating \reff{integrate} from $0$ to $t(\leq\hat{t}+\delta)$, we obtain \begin{equation*}
\|x_{t}-\hat{x}_{t}\|^{2}\leq 2L_\xi\int_{0}^{t}\|x_{s}-\hat{x}_{s}\|^{2}ds.
\end{equation*} Applying Gronwall's inequality\citep{Aubin1984Differential} to the above inequality, we have that $x_t=\hat{x}_t,\forall{t}\in[0,\hat{t}+\delta]$, which leads to a contradiction. Therefore, the solution of \reff{Neural Network} with $x_0\in\mathcal{X}$ is unique.
	
Finally, we prove the global Lipschitz continuity of $\dot{x}_t$ on $[0,+\infty)$. By the definition of $\mu_t$ in \reff{Neural Network}, $\dot{\mu}_t$ is bounded and globally Lipschitz continuous on $[0,+\infty)$.
From Proposition \ref{prop1}-(iii) and the boundedness of $\dot{x}_t$ and ${\mu}'_t$ on $[0,+\infty)$, since $x_t\in\mathcal{X}$ and $\mu_t\in[\frac{1}{2}\mu^*,\mu_0]$ for all $t\in[0,+\infty)$, we obtain the global Lipshictz continuity of $\nabla_x\Theta(x_t,\mu_t)$
on $[0,+\infty)$. The global Lipschitz continuity of $\nabla f(x)$ on $\mathcal{X}$ and the boundedness of $\dot{x}_t$ on $[0,+\infty)$ implies the global Lipschitz continuity of $\nabla f(x_t)$ on $[0,+\infty)$. Combining the above analysis, the global Lipschitz continuity of projection operator $P_{\mathcal{X}}$ given in Proposition \ref{projection operator property} and
the structure of $\dot{x}_t$ in \reff{Neural Network}, we conclude that $\dot{x}_t$ is global Lipschitz continuous on $[0,+\infty)$.
\end{pf}

The following result plays an important role in the convergence analysis of neural network \reff{Neural Network} for solving \reff{optimization model}, which gives some basic dynamic properties of the solution of \reff{Neural Network}.

\begin{lemma}\label{lemma1}
Suppose $x_t$ is the solution of neural network \reff{Neural Network} with initial point $x_{0}\in \mathcal{X}$, then we have
\begin{enumerate}[i)]
\item $\lim_{t\rightarrow+\infty}\left[f(x_{t})+\lambda\Theta\left(x_{t}, \mu_t\right)\right]$ exists;
\item $\int_0^{+\infty}\|\dot{x}_{t}\|^2 dt<+\infty$ and $\lim_{t\rightarrow+\infty}\dot{x}_{t}=\bm{0}$.
\end{enumerate}
\end{lemma}
\begin{pf}
See Appendix \ref{a2}.
\end{pf}
\begin{remark}
Based on the boundedness of $x(t)$ on $[0,\infty)$ and $\lim_{t\rightarrow+\infty}\dot{x}_t=\bm{0}$, we have that any accumulation point of $x(t)$ is a ${\mu^*}/{2}$-stationary point by Definition \ref{stationary} of \reff{smoothing optimization model}.
\end{remark}
\subsection{Properties of the accumulation points of the solution to network \reff{Neural Network}}\label{section3.2}
In this subsection, we analyze the optimal properties of the solution to network \reff{Neural Network} for problem \reff{optimization model}, which lays a foundation for the effectiveness of the proposed method in this paper for solving \reff{optimization model}. This piecewise property of $\Theta$ is the key to establishing the relationships between the accumulation points of (\ref{Neural Network}) and the optimal solutions of (\ref{optimization model}).

For the further analysis on (\ref{Neural Network}), let us give some necessary notations.
For an $x\in \mathcal{X}$, define \begin{equation*}
I(x)=\left\{i\in\{1,2,\ldots,n\}: x_i\in\left[0,\frac{1}{6}\mu^*\right)\right\},
\end{equation*} \begin{equation*}
{J(x)}=\left\{j\in\{1,2,\ldots,n\}:x_j\in\left[\frac{1}{6}\mu^*,\frac{1}{2}\mu^*\right)\right\}
\end{equation*} and \begin{equation*}
K(x)=\left\{k\in\{1,2,\ldots,n\}:x_k\in\left[\frac{1}{2}\mu^*,\upsilon_k\right]\right\}.
\end{equation*} It is clear that $I(x)\cup J(x) \cup K(x)=\{1,2,\ldots,n\}$.

For set $\mathcal{X}$ in \reff{optimization model} and an $x\in\mathcal{X}$, denote $\mathcal{X}_{I}(x)=\{y\in\mathcal{X}: y_i=0, \forall i\in {I(x)}\}$, $\mathcal{X}_{J}(x)=\{y\in \mathcal{X}: y_i=x_i, \forall i\in {J(x)}\}$ and $\mathcal{X}_{K}(x)=\{y\in \mathcal{X}: y_i=x_i, \forall i\in {K(x)}\}$. Let $\mathcal{X}_{I\cap J}(x)=\mathcal{X}_{I}(x)\cap\mathcal{X}_{J}(x)$.

To show the good performance of the proposed network in solving problem \reff{optimization model}, we first prove a unified lower bound property and the same support set property for the accumulation points of the solution to network \reff{Neural Network}.	

\begin{lemma}\label{lemma2}
Let $x_t$ be the solution of neural network \reff{Neural Network} with initial point $x_{0}\in\mathcal{X}$ and suppose that $\bar{x}$ be an accumulation point\footnote{$\bar{x}$ is called an accumulation point of the solution $x(t)$ to (\ref{Neural Network}), if there exists a sequence $\{t_k\}$ with $\lim_{k\rightarrow+\infty}t_k=+\infty$, such that $\lim_{k\rightarrow+\infty}x(t_k)=\bar{x}$.} of $x_t$, then,
\begin{equation}\label{lower bound}
\mbox{either $\bar{x}_i=0$ or $\bar{x}_i\geq\frac{1}{6}\mu^*$, for $i=1,\ldots,n$,}
\end{equation}
and hence $I(\bar{x})=\{i\in\{1,2,\ldots,n\}:\bar{x}_i=0\}$.
Moreover, for any accumulation points $\hat{x}$ and $\tilde{x}$ of $x_{t}$, it holds that $I(\hat{x})=I(\tilde{x})$.
\end{lemma}
\begin{pf}
See Appendix \ref{a3}.
\end{pf}

For algorithms that solve sparse regression problems, the results in Lemma \ref{lemma2} are very important for the numerical properties of the proposed method. There are some analysis on the lower bound property for many different sparse regression models \citep{Chen2012Non,Bian2015Optimality,Chartrand2008Restricted}. However, most of the results are proved for the local minimizer. The first result in Lemma \ref{lemma2} indicates that all accumulation points of the solution of network \reff{Neural Network} have a unified lower bound for nonzero entries. The lower bound property of the accumulation points shows that network \reff{Neural Network} can distinguish zero and nonzero entries of coefficients effectively in sparse high-dimensional regression \citep{Chartrand2008Restricted, Huang2008Asymptotic}, and bring the restored image closed contours and neat edges \citep{ Chen2012Non}. Moreover, it is worth noting that the lower bound is related to the regularization parameter $\lambda$. Wherefore, the lower bound is useful for choosing the regularization parameter $\lambda$ to control the sparsity of the accumulation points of network \reff{Neural Network}. Through the accumulation points of network \reff{Neural Network} may be not unique, the second result in Lemma \ref{lemma2} shows that all accumulation points own a common support set, which shows the constancy and robustness of network \reff{Neural Network} for solving problem \reff{optimization model}.

Next, we give a sufficient and necessary optimality condition for the local minimizers of (\ref{optimization model}), which helps us to justify the optimal property of the obtained point. Based on the special discontinuity of $\|x\|_0$ and the continuity of $f$, we have the following link on the local minimizers of $f(x)+\lambda\|x\|_0$ and $f(x)$.
\begin{proposition}\label{==}
$x^*$ is a local minimizer of (\ref{optimization model}) if and only if $x^*$ is a local minimizer of $f$ in $\mathcal{X}_{I}(x^*)$.
\end{proposition}
\begin{pf}
See Appendix \ref{a4}.
\end{pf}

Based on the properties proved in Lemma \ref{lemma2} and Assumption \ref{choice}, any accumulation point of the solution to network \reff{Neural Network} owns the following optimal properties to problem \reff{optimization model}.

\begin{theorem}\label{empty}
Let $x_t$ be the solution of neural network \reff{Neural Network} with initial point $x_{0}\in\mathcal{X}$ and suppose $\bar{x}$ be an accumulation point of $x_t$. Then, $\bar{x}$ is a global minimizer of $f(x)$ in $\mathcal{X}_{I\cap J}(\bar{x})$. In particular, if $J(\bar{x})=\emptyset$, then $\bar{x}$ is a local minimizer of sparse regression model \reff{optimization model} with lower bound property in \reff{lower bound}.
\end{theorem}
\begin{pf}
Taking into account that $\bar{x}_k\geq\frac{1}{2}\mu^*,\forall k\in{K(\bar{x})}$, by \reff{Respectively Derivative x2}, we obtain
\begin{align}\label{K0}
\left[\nabla_x\Theta\left(\bar{x},{\mu^*}/{2}\right)\right]_k={0},\forall k\in{K(\bar{x})}.
\end{align}Denote $\mathcal{X}_{K(\bar{x})}:=\{x_{K(\bar{x})}:x\in\mathcal{X}\}$. Recalling $\lim_{t\rightarrow+\infty}\dot{x}_t=\bm{0}$ and applying \reff{K0}, since feasible set $\mathcal{X}$ is box shaped, we have
\begin{align*}
\begin{split}
\bar{x}_{K(\bar{x})}&=P_{\mathcal{X}_{K(\bar{x})}}\left[\left[\bar{x}-\nabla{f}(\bar{x})-\lambda\nabla_x\Theta\left(\bar{x},{\mu^*}/{2}\right)\right]_{K(\bar{x})}\right]\\&=P_{\mathcal{X}_{K(\bar{x})}}\left[\bar{x}_{K(\bar{x})}-\left[\nabla{f}(\bar{x})\right]_{K(\bar{x})}\right].
\end{split}
\end{align*}
Based on Proposition \ref{minimizer}, $\bar{x}$ is a global minimizer of $f$ in $\mathcal{X}_{I\cap J}(\bar{x})$.
	
If $J(\bar{x})=\emptyset$, by Proposition \ref{==}, $\bar{x}$ is a local minimizer of $f(x)+\lambda\|x\|_0$ in $\mathcal{X}$ and satisfies the lower bound property in \reff{lower bound}.
\end{pf}
\begin{remark}\label{sta}
By the proofs of Lemma \ref{lemma2} and Theorem \ref{empty}, for the stationary point $x^s_{\mu^*/2}$ of \reff{smoothing optimization model}, if $J(x^s_{\mu^*/2})=\emptyset$, then $x^s_{\mu^*/2}$ is a local minimizer of \reff{optimization model}.
\end{remark}

Due to the special structure of function $\theta(\cdot,\mu)$, which is convex on $[0,\frac{1}{3}\mu]$ and $[\mu,+\infty)$, respectively, the solution of network \reff{Neural Network} can be convergent to a local minimizer of \reff{optimization model} in some cases.
\begin{corollary}
Let $x_t$ be the solution of \reff{Neural Network}. Suppose that $f$ in \reff{optimization model} is strictly convex and any accumulation point $\bar{x}$ of $x_t$ satisfies $J(\bar{x})=\emptyset$. Then $x_{t}$ is convergent to a local minimizer of problem \reff{optimization model} with lower bound property in \reff{lower bound}.
\end{corollary}

\subsection{A further correction to the obtained point by \reff{Neural Network}}\label{section5}
In this subsection, we propose a further network to correct the obtained accumulation point of the solution to \reff{Neural Network}. Its solution can converge to a local minimizer of problem \reff{optimization model}. Moreover, in the sense of the sparsity and objective function value, it can find a better solution than the accumulation point obtained by \reff{Neural Network}. Throughout this subsection, denote $\bar{x}$ as an accumulation point of the solution to \reff{Neural Network}.

The proposed network in this part should be with a special initial point, which is constructed by $\bar{x}$ and defined as follows.
\begin{definition}
We call $\bar{x}^{\mu^*}$ a $\mu^*$-update point of $\bar{x}$, if
\begin{gather}\label{update}
\bar{x}^{\mu^*}_i=\left\{
\begin{aligned}
&\bar{x}_i~~~&&{\rm if}~|\bar{x}_i|\geq\frac{\mu^*}{2},\\
&0~~~~~&&\rm{otherwise}.
\end{aligned}
\right.
\end{gather}
\end{definition}
Obviously, $\bar{x}=\bar{x}^{\mu^*}$ if $J(\bar{x})=\emptyset$.

Though the $\mu^*$-update point of $\bar{x}$ is also not a local minimizer of \reff{optimization model}, it is a local minimizer of $f(x)+\lambda\|x\|_0$ in a particular subset of $\mathcal{X}$ and owns some special properties.
\begin{proposition}\label{accumulation}
The $\mu^*$-update point $\bar{x}^{\mu^*}$ of $\bar{x}$ is a strictly local minimizer of $f(x)+\lambda\|x\|_0$ in $\mathcal{X}_{K}(\bar{x})$ and $f(\bar{x}^{\mu^*})+\lambda\|\bar{x}^{\mu^*}\|_0\leq f(\bar{x})+\lambda\|\bar{x}\|_0$. In particular, if $J(\bar{x})\neq\emptyset$, then $f(\bar{x}^{\mu^*})+\lambda\|\bar{x}^{\mu^*}\|_0< f(\bar{x})+\lambda\|\bar{x}\|_0$.
\end{proposition}
\begin{pf}
See Appendix \ref{a5}.
\end{pf}

Denote $\bar{\mathcal{X}}=\mathcal{X}_I(\bar{x}^{\mu^*})$,
which is the set defined by $\bar{x}$, i.e. $\bar{\mathcal{X}}=\{x\in\mathcal{X}:x_i=0,\forall i\in I(\bar{x})\cup J(\bar{x})\}$.

We propose a further network for the case of $J(\bar{x})\neq\emptyset$. The further network is a projection neural network modeled by the following differential equation:
\begin{equation}\label{Neural Network3}
\left\{
\begin{split}	&\dot{{x}}(t)=\gamma_1\Big[-{x}(t)+P_{\bar{\mathcal{X}}}\left[{x}(t)-{\nabla{{f}}\left({x}(t)\right)}\right]\Big]\\
&{x}(0)=\bar{x}^{\mu^*},
\end{split}
\right.
\end{equation}
where $\gamma_1$ is a given positive parameter. Substantially, network \reff{Neural Network3} is used to solve the following smooth convex optimization problem:
\begin{equation}\label{sub-problem}
\min_{x\in\bar{\mathcal{X}}}\quad f(x).
\end{equation}

\begin{remark}\label{special}
If $J(\bar{x})=\emptyset$, then $\bar{x}$ is a local minimizer of problem \reff{optimization model} with lower bound property by Theorem \ref{empty} and an equilibrium point of its corresponding correction method. Therefore, for the case of $J(\bar{x})=\emptyset$, the correction method does not need to be used.
\end{remark}
\begin{remark}
Though the variable $x(t)$ in network \reff{Neural Network3} is with dimension $n$, by the uniqueness of the value of the points in set $\bar{\mathcal{X}}$ for the index in $I(\bar{x}^{\mu^*})$, the calculation dimension of $x(t)$ in \reff{Neural Network3} is $n-|I(\bar{x}^{\mu^*})|$.
\end{remark}

Since network \reff{Neural Network3} is a typical projection neural network for a constrained smooth convex optimization problem, the solution of \reff{Neural Network3} is global existent, unique and owns the following properties \citep{Li2010Generalized}:
\begin{itemize}
\item ${x}_t\in\bar{\mathcal{X}}$;
\item ${f}({x}_{t})$ is nonincreasing in $t$ and $\lim_{t\rightarrow+\infty}{f}({x}_{t})$ exists;
\item $\lim_{t\rightarrow+\infty}\dot{{x}}_{t}=\bm{0}$;
\item $x_t$ is convergent to a global minimizer of problem \reff{sub-problem}.
\end{itemize}

Based on the above basic properties of network \reff{Neural Network3} and its particular initial point, we obtain the following result on the limit point of its solution to problem \reff{optimization model}.
\begin{theorem}\label{corre}
Denote $x^*$ the limit point of the solution to network \reff{Neural Network3} with initial point $\bar{x}^{\mu^*}$. Then, $x^*$ is a local minimizer of \reff{optimization model}, and if $J(\bar{x})\neq\emptyset$, the following properties hold:
\begin{enumerate}[i)]
\item $\|x^*\|_0< \|\bar{x}\|_0$;
\item $f(x^*)+\lambda\|x^*\|_0< f(\bar{x})+\lambda\|\bar{x}\|_0$.
\end{enumerate}
\end{theorem}
\begin{pf}
Denote $x_t$ the solution of network \reff{Neural Network3} with initial point $\bar{x}^{\mu^*}$. By $\lim_{t\rightarrow+\infty}\dot{{x}}_{t}=\bm{0}$, we have \begin{equation*}
{x}^*=P_{\bar{\mathcal{X}}}\left[{x}^*-{\nabla{{f}}\left({x}^*\right)}\right].
\end{equation*}
Since $x_t\in\bar{\mathcal{X}}$, we have ${x}^*\in\mathcal{X}_{I}({x}^*)\subseteq \bar{\mathcal{X}}$, and hence \begin{equation*}
{x}^*=P_{{\mathcal{X}}_{I}({x}^*)}\left[{x}^*-{\nabla{{f}}\left({x}^*\right)}\right].
\end{equation*}
By Proposition \ref{minimizer}, we obtain that ${x}^*$ is a global minimizer point of ${f}(x)$ in $\mathcal{X}_{I}({x}^*)$. Based on Proposition \ref{==}, we know that ${x}^*$ is a local minimizer point of problem \reff{optimization model}. When $J(x)\neq\emptyset$, by $I(\bar{x})\subsetneq I(\bar{x}^{\mu^*})\subseteq {I}({x}^*)$ and the nonincreasing property of $f(x_t)$ in $t$, we have $\|\bar{x}\|_0>\|\bar{x}^{\mu^*}\|_0\geq\|{x}^*\|_0$ and $f(\bar{x}^{\mu^*})\geq f(x^*)$. By Proposition \ref{accumulation}, if $J(\bar{x})\neq\emptyset$, then $f({x}^*)+\lambda\|{x}^*\|_0\leq f(\bar{x}^{\mu^*})+\lambda\|\bar{x}^{\mu^*}\|_0< f(\bar{x})+\lambda\|\bar{x}\|_0$.
\end{pf}
\begin{remark}
By Theorem \ref{empty} and Theorem \ref{corre}, either the accumulation point $\bar{x}$ of \reff{Neural Network} satisfying $J(\bar{x})=\emptyset$ or the accumulation point $x^*$ of \reff{Neural Network3} with $J(\bar{x})\neq\emptyset$, is a local minimizer of \reff{optimization model}.
\end{remark}
\section{Extension to another regression model}\label{section6}
In this section, we consider another sparse regression model
\begin{equation}\label{two side}
\begin{split}
&\min\quad f(y)+\lambda\|y\|_0\\
&~\text{s.t.}~\quad y\in{\mathcal{Y}}:=\{y\in\mathbb{R}^{n}:-l\leq{y}\leq u\},
\end{split}
\end{equation}
where $l,\,u\in\mathbb{R}^n_+$, $\lambda>0$, $f$ is defined as in \reff{optimization model}.

To solve \reff{two side}, we consider an sparse regression model modeled by \reff{optimization model} with $2n$ dimension, i.e.
\begin{equation}\label{one side}
\begin{aligned}
&\min\quad &&f(x_+-x_-)+\lambda\|x_+\|_0+\lambda\|x_-\|_0\\
&~\text{s.t.}~\quad &&x_+\in {\mathcal{X}_1}:=\{x_+\in\mathbb{R}^{n}:{\bm 0}\leq{x_+}\leq u\},\\
&\,&&x_-\in{\mathcal{X}_2}:=\{x_-\in\mathbb{R}^{n}:\bm{0}\leq{x_-}\leq l\}.
\end{aligned}
\end{equation}

The following proposition indicates the link between sparse regression models \reff{two side} and \reff{one side}.
\begin{proposition}\label{xx+x-}
Optimization problems \reff{two side} and \reff{one side} are equivalent and their local minimizers can be converted to each other, i.e. for any local minimizer $y^*$ of \reff{two side}, there exists a local minimizer $({x^*_+}^\top , {x^*_-}^\top)^\top$ of \reff{one side} such that $y^*=x^*_+ -x^*_-$ and vice versa.
\end{proposition}
\begin{pf}
See Appendix \ref{a6}.
\end{pf}
\begin{remark}
Proposition \ref{xx+x-} shows that the given method in this paper can also be used to solve \reff{two side0} by solving \reff{oneside0}. It is worth noting that it is impossible to solve \reff{two side0} directly without \reff{oneside0} by differential equations in theory. In fact, $\Theta$ is a smoothing function of $\|x\|_0$ on $\mathbb{R}^n_+$, but not $\mathbb{R}^n$. Based on the construction of $\Theta$, it is easier to have a similar smoothing approximation function of $\|x\|_0$ on $\mathbb{R}^n$. However, it is not continuously differentiable everywhere, which involves subdifferential in the proposed network. Neural network \reff{Neural Network} is modeled by a differential equation. It is more conducive to solving \reff{optimization model} than differential inclusion in computation and implementation.
\end{remark}
\section{Numerical experiments}\label{section7}
In this section, we report four numerical experiments to validate the theoretical results and show the efficiency of neural network \reff{Neural Network} for solving \reff{optimization model} and \reff{two side}. Let MSE$(x)=\|x-s\|^2/n$ denote the mean squared error of $x$ to original signal $s\in\mathbb{R}^n$. The average, maximum and minimum mean square error of test results are denoted by Mean-MSE, Max-MSE and Min-MSE, respectively. The average, maximum and minimum CPU time of test results are denoted by Mean-CPU, Max-CPU and Min-CPU, respectively.

For function $f(x)=\|Ax-b\|^2$, where $A\in\mathbb{R}^{m\times n}$ and $ b\in\mathbb{R}^m$, we use the following code to generate $L_f$ ($\rm{L}$ in code), which is an upper bound of \begin{equation*}
\left\{||\nabla f(x)||_\infty: x\in[\bm{0},\textbf{k}_n]\right\},
\end{equation*} for given $k>0$.
\begin{lstlisting}
[m,n]=size(A); W=abs(A);
C1=2*(W'*W*k*ones(n,1)-A'*b);
C2=2*(-W'*W*k*ones(n,1)-A'*b);
L1=norm(C1,inf);L2=norm(C2,inf);
L=max([L1 L2]);
\end{lstlisting}
Specially, if all entries of $A$ and $b$ are nonnegative, then we can make $\rm{L}=\rm{L1}$. Further, for general sparse regression problem \reff{optimization model}, we let \begin{equation*}
\mu^*=0.9\min\left\{k,\frac{3\lambda}{2(k+L_f)},\frac{2\lambda}{{n}L_f}\right\}.
\end{equation*}

We should state that the correction method in Subsection \ref{section5} is not used in all numerical experiments, since the limit point $x^*$ of the solution of neural netwrok \reff{Neural Network} satisfies $J(x^*)=\emptyset$ in every experiment.
\subsection{Test example}\label{test1}
In this example, we illustrate the effectiveness of the algorithm for solving a test problem modeled by \reff{optimization model}, whose global minimizers are known.

Consider the following sparse regression problem in $\mathbb{R}^2$:
\begin{equation}\label{test}
\begin{split}
&\min~\|Ax-b\|^2+\lambda\|x\|_0\\
&~\text{s.t. } ~ x\in\mathcal{X}:=\{x\in\mathbb{R}^{2}:\bm{0}\leq{x}\leq\upsilon\},
\end{split}
\end{equation}
where
\begin{equation}
A=
\left(
\begin{array}{ccc}
1& 3 &1\\
3& 2 &5\\
\end{array}
\right)^\top,
\end{equation}
$b=(2,1,3)^\top$, $\lambda=1$ and $\upsilon=(5,5)^\top$.

\begin{figure}
\centering
\subfigure[]{\includegraphics[width=2.5in]{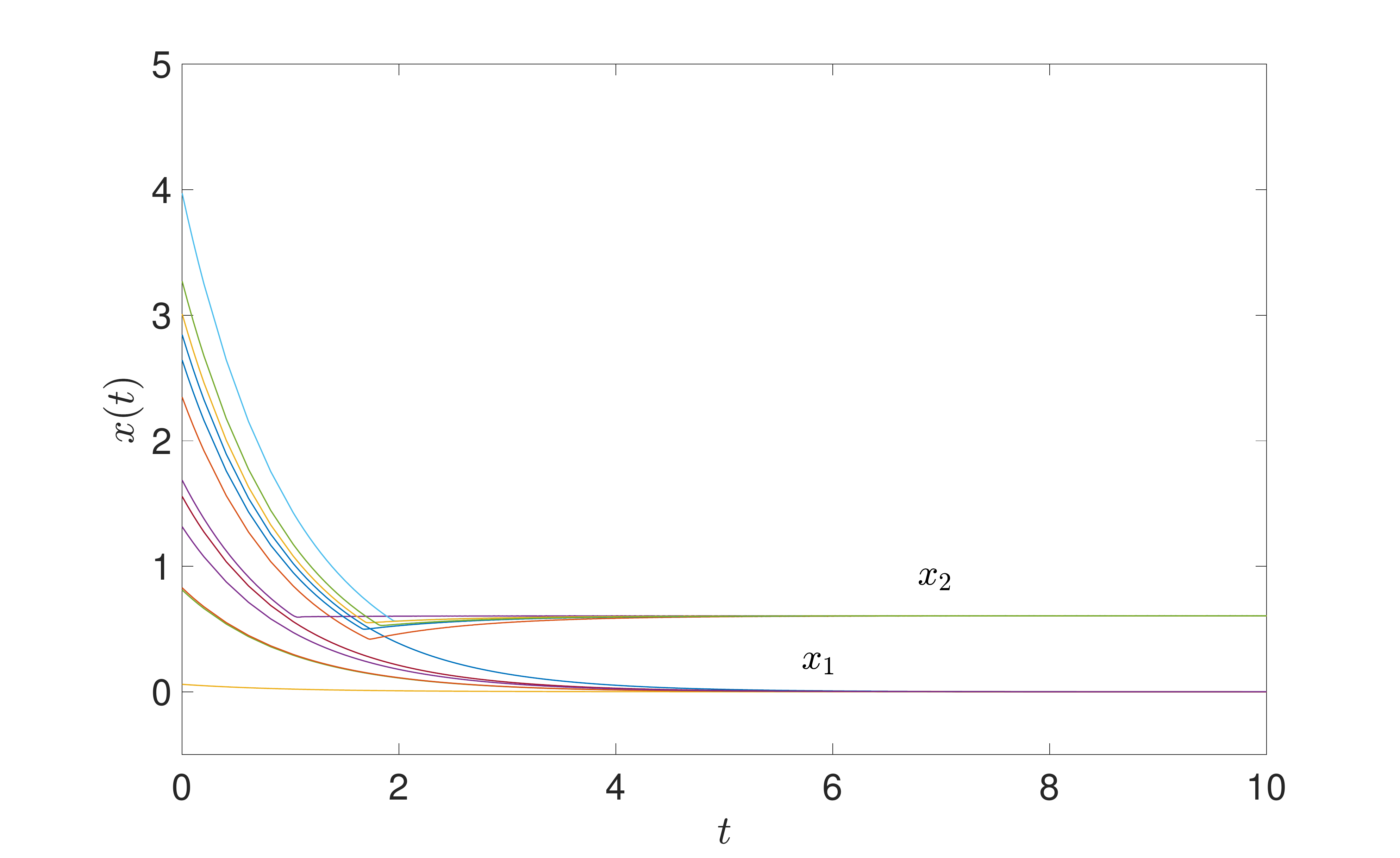}
\label{fig:variable}}
\hfil
\subfigure[]{\includegraphics[width=2.5in]{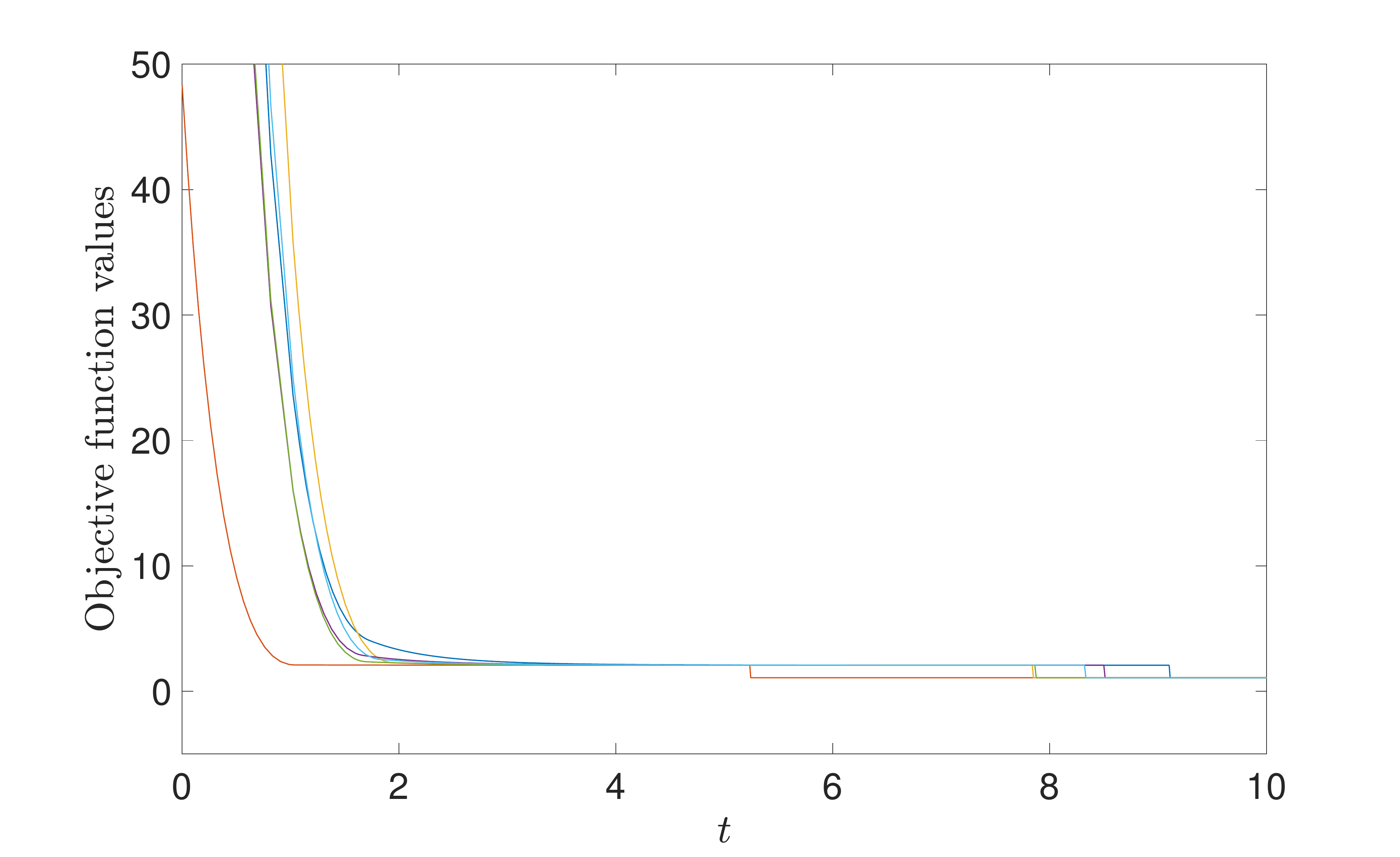}
\label{fig:value}}
\caption{(a) State trajectories and (b) objective function values along the solutions of neural network \reff{Neural Network} for problem \reff{test}.} \label{fig:-}
\end{figure}

We implement network \reff{Neural Network} with $\alpha_0=\beta=\gamma=1$ and stop when $t=10$.  Fig. \ref{fig:-} (a) presents the solution of \reff{Neural Network} with $6$ random initial points in $\mathcal{X}$, which converge to the global minimizer point $x^*=(0,0.6053)^\top$ of \reff{test}. Fig. \ref{fig:-} (b) shows that the objective function values are decrease along the solutions of neural network \reff{Neural Network} with the same $6$ initial points used in Fig. \ref{fig:-} (a).
\subsection{Compressed sensing}\label{Signal}
To validate Proposition \ref{xx+x-}, we consider the following constrained sparse regression problem:
\begin{equation}\label{compressed sensing}
\begin{split}
&\min~\|Ax-b\|^2+\lambda\|x\|_0\\
&~\text{s.t. } ~ x\in\mathcal{X}:=\{x\in\mathbb{R}^{n}:-\upsilon\leq{x}\leq\upsilon\},
\end{split}
\end{equation}
where $A\in\mathbb{R}^{m\times n}$, $b\in\mathbb{R}^m$, $n=1000$, $m=200$, $\lambda=0.1$ and $\upsilon=\textbf{5}_n$. Sensing matrix $A$, original signal $s$ with $\|s\|_0=10$ and observation $b$ are generated by the following code:
\begin{lstlisting}
K=randn(n,m);A'=orth(K);
Q=randperm(n);s=zeros(n,1);
s(Q(1:10))=randn(10,1);b=A*s;
\end{lstlisting}
By Proposition \ref{xx+x-}, sparse regression problem \reff{compressed sensing} is equivalent to the following sparse regression problem:
\begin{equation}\label{transform}
\begin{split}
&\min~\|Ax_+-Ax_--b\|^2+\lambda\|x_+\|_0+\lambda\|x_-\|_0\\
&~\text{s.t.}~x_+,x_-\in\mathbb{R}^{n}, \bm{0}\leq{x_+,x_-}\leq\upsilon.
\end{split}
\end{equation}

\begin{figure}
\centering
\subfigure[]{\includegraphics[width=2.5in]{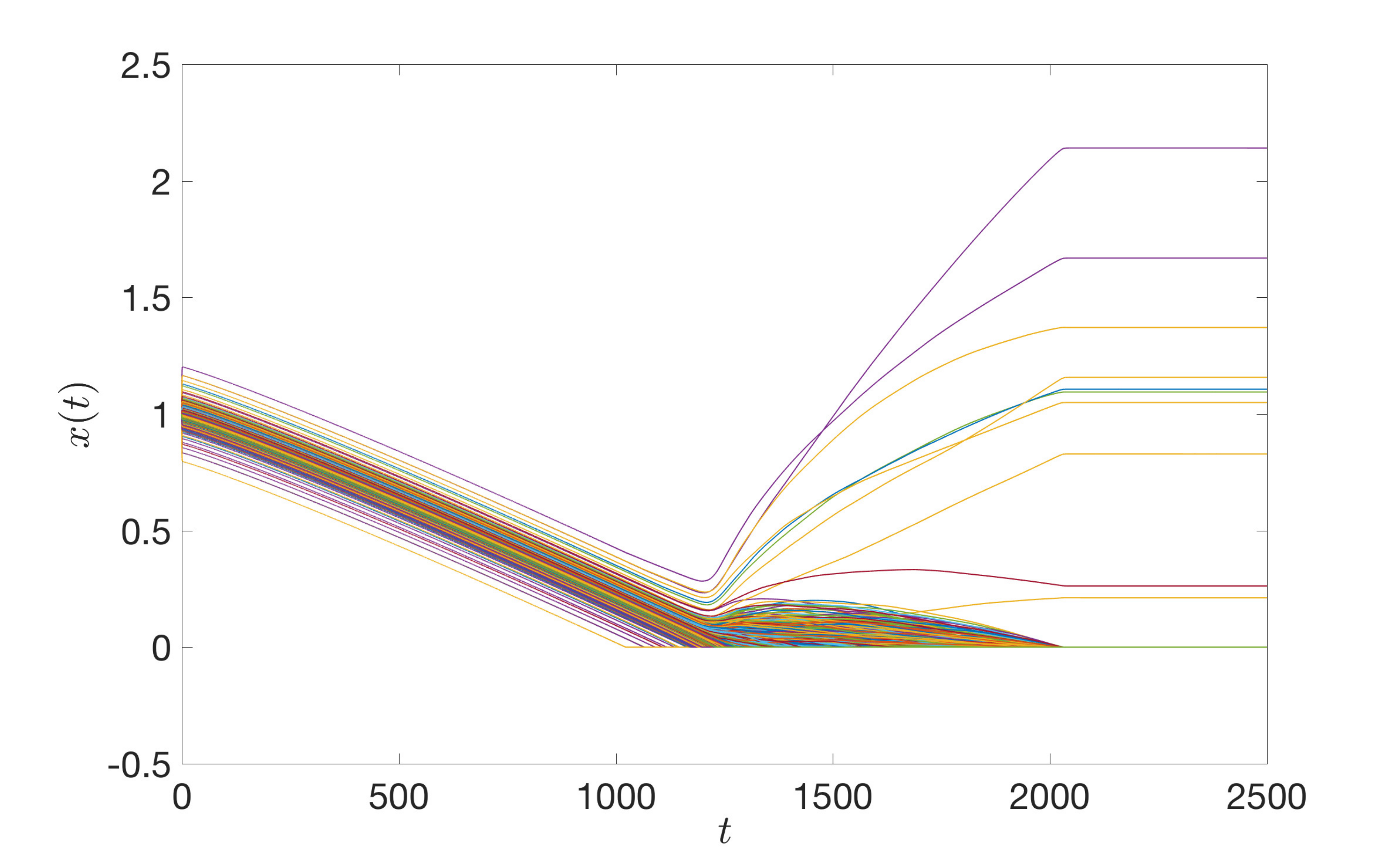}
\label{x}}
\hfil
\subfigure[]{\includegraphics[width=2.5in]{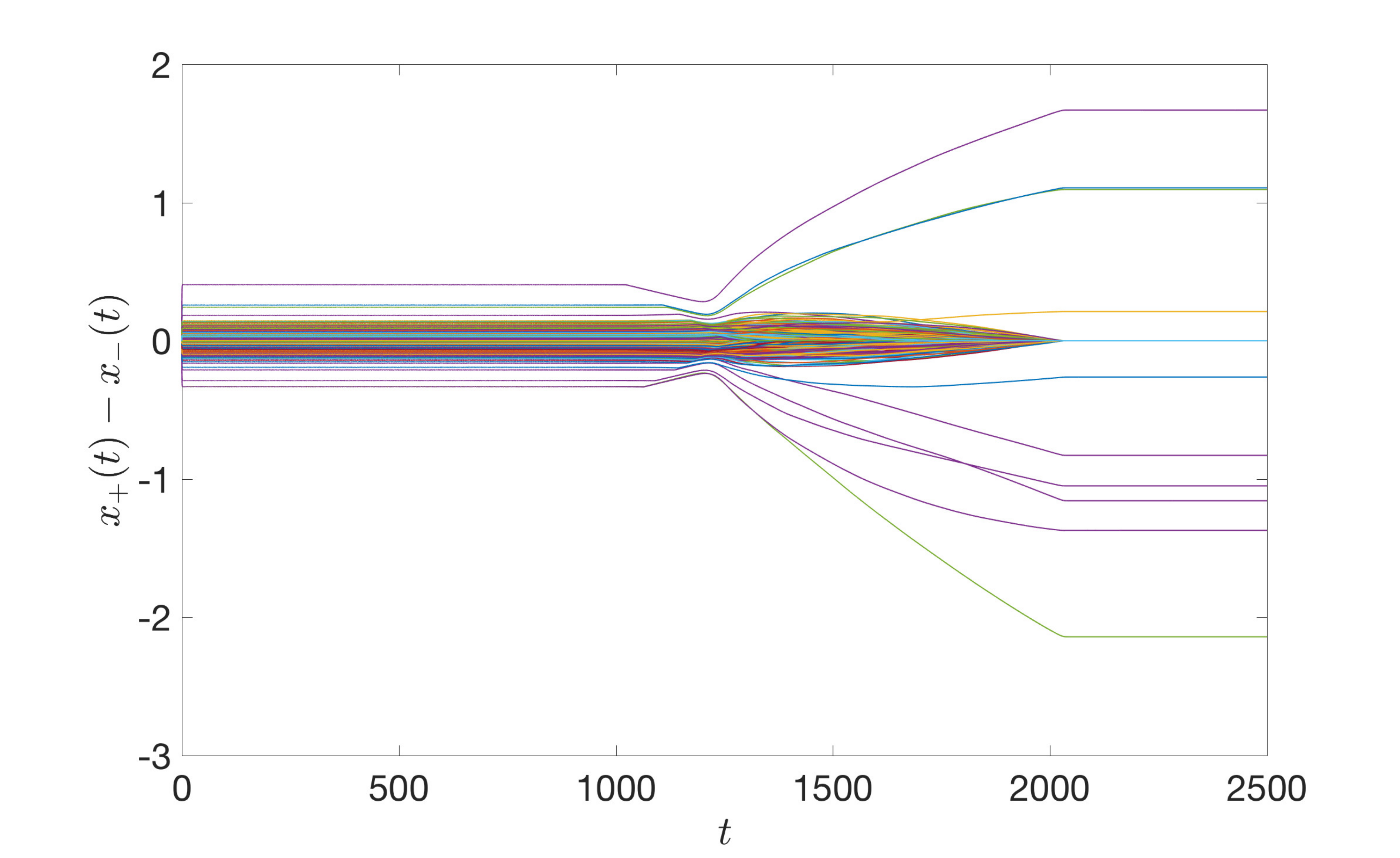}
\label{x_+-x_-}}
\caption{(a) State trajectory $x(t)$ and (b) $x_+(t)-x_-(t)$ of neural network \reff{Neural Network} for problem \reff{compressed sensing}.} \label{fig:x}
\end{figure}

\begin{figure}
\centering
\includegraphics[width=2.5in]{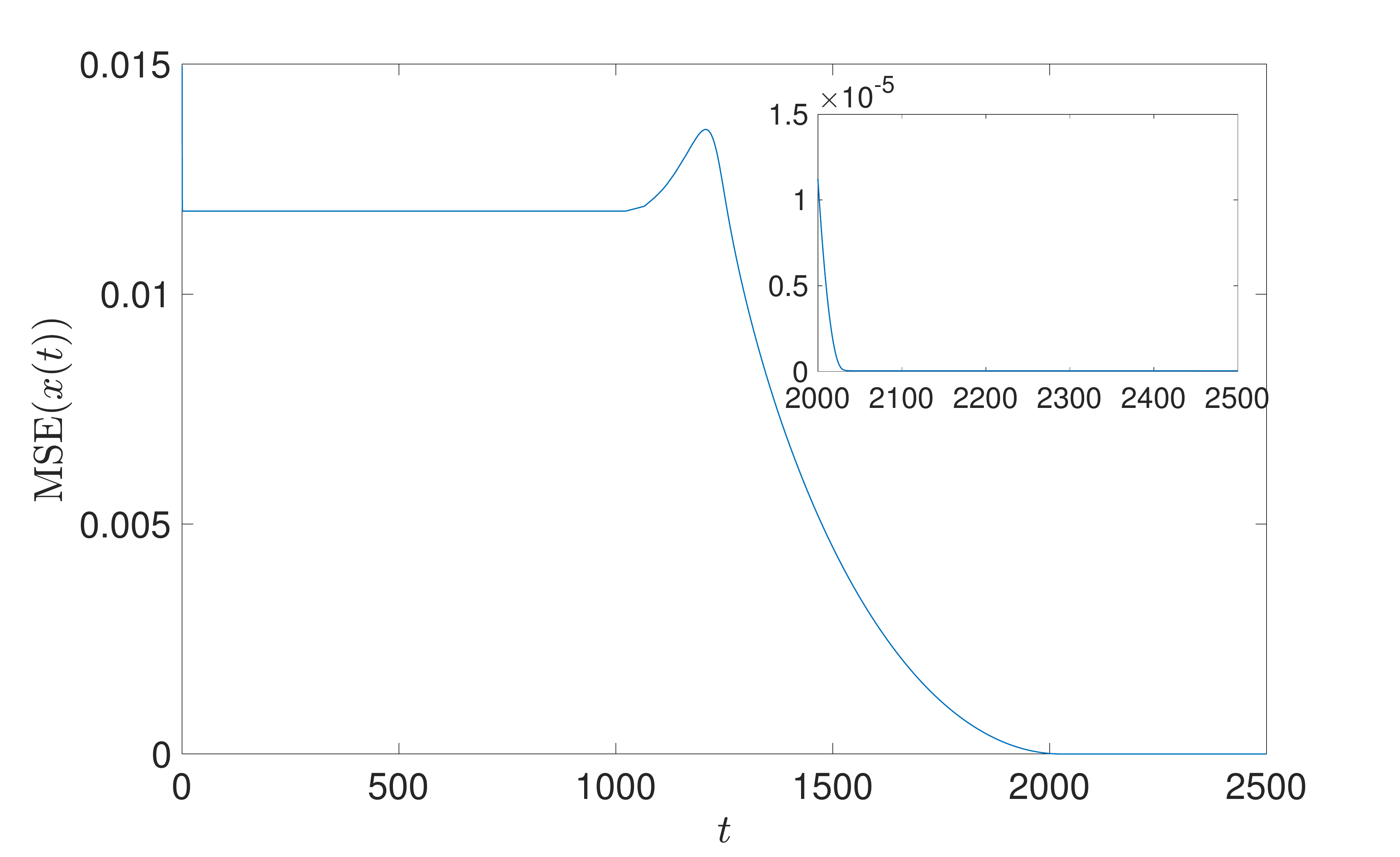}
\caption{Mean squared error of $x_+(t)-x_-(t)$ to the original signal.}
\label{fig:M}
\end{figure}

\begin{figure}
\centering
\includegraphics[width=2.5in]{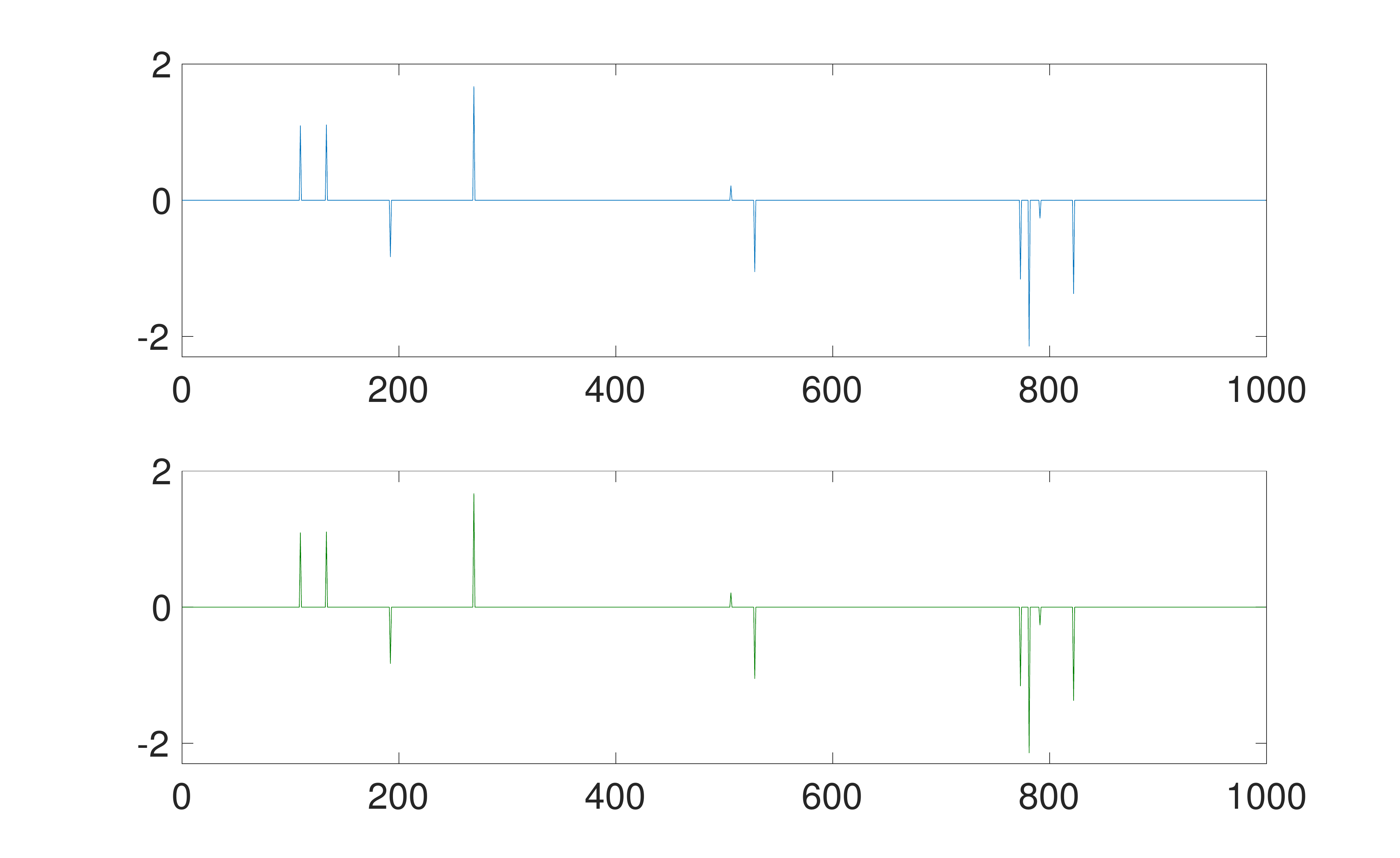}
\caption{Original signal and reconstructed signal $x_+(t)-x_-(t)$ by neural network \reff{Neural Network}, where the one above is original signal.} \label{fig:S}
\end{figure}

Choose $\alpha_0=700$, $\beta=0.1$, $\gamma=1$, ten random initial points in this set $[0,1]^{2n}$ and a fixed initial point $x_0=\textbf{1}_{2n}$ in neural network (\ref{Neural Network}) to solve problem (\ref{transform}). We implement network (\ref{Neural Network}) and stop when $t=2500$. Fig. \ref{fig:x} (a) shows that solution $x(t)$ of \reff{Neural Network} with $x_0=\textbf{1}_{2n}$ is convergent. Fig. \ref{fig:x} (b) gives the transformation $x_+(t)-x_-(t)$ of solution $x(t)$ with $x_0=\textbf{1}_{2n}$, which is used for solving sparse regression problem \reff{compressed sensing}. Fig. \ref{fig:M} presents the mean squared error of $x_+(t)-x_-(t)$ to original signal $s$ with respect to $t$ in neural network \reff{Neural Network} with $x_0=\textbf{1}_{2n}$. Fig. \ref{fig:S} shows the original and reconstructed signals by neural network \reff{Neural Network} with $x_0=\textbf{1}_{2n}$, where the one above is the original signal. We can see that the original signals almost coincide with the reconstructed signals. In Table \ref{ten}, MSE* means the mean squared error of the output solution of neural network \reff{Neural Network} with $x_0=\textbf{1}_{2n}$ and other data are the mean squared errors of the output solution of neural network \reff{Neural Network} with ten random initial points, which are very close to the result MSE* with $x_0=\textbf{1}_{2n}$. Therefore neural network \reff{Neural Network} is insensitive to initializations and we use the fixed initial point of all ones in the following experiments.

\begin{table}[!]
\renewcommand{\arraystretch}{1.3}
\caption{MSEs with Neural Network \reff{Neural Network} with ten random initial points and the fixed initial point $\textbf{1}_{2n}$}
\label{ten}
\centering \begin{tabular}{|c|c|c|c|}
\hline
Mean-MSE& Max-MSE & Min-MSE & MSE* \\
\hline
$4.2049\times10^{-8}$& $4.2050\times10^{-8}$ & $4.2048\times10^{-8}$ &$4.2050\times10^{-8}$\\
\hline
\end{tabular}
\end{table}
\subsection{Variable selection}
Variable selection is an important application in high-dimensional statistical problems, particularly in regression and classification problems. We consider this problem by the following sparse regression model:
\begin{equation}\label{regression}
\begin{split}
&\min~\|Ax-b\|^2+\lambda\|x\|_0\\
&~\text{s.t. } ~ x\in\mathcal{X}:=\{x\in\mathbb{R}^{n}:\bm{0}\leq{x}\leq\upsilon\},
\end{split}
\end{equation}
where $A\in\mathbb{R}^{m\times n}$, $b\in\mathbb{R}^m$, $\lambda=1$ and $\upsilon\in\mathbb{R}_+^n$.

Let $n=1500$, $m=600$ and $\upsilon=\textbf{10}_n$. We use the following code to generate measurement matrix $A$, observation $b$ and original signal $s$ with $\|s\|_0=50$:
\begin{lstlisting}
L=randn(n,m);A'=orth(L);
P=randperm(n);s=zeros(n,1);
s(P(1:50))=unifrnd(1,10,[50,1]);
b=A*s+0.01*randn(m,1);
\end{lstlisting}

In this example, we  consider some comparative experiments and implement network \reff{Neural Network} with the same parameters and initial point as in example 6.2.

\begin{figure}[!t]
\centering
\subfigure[]{\includegraphics[width=1in]{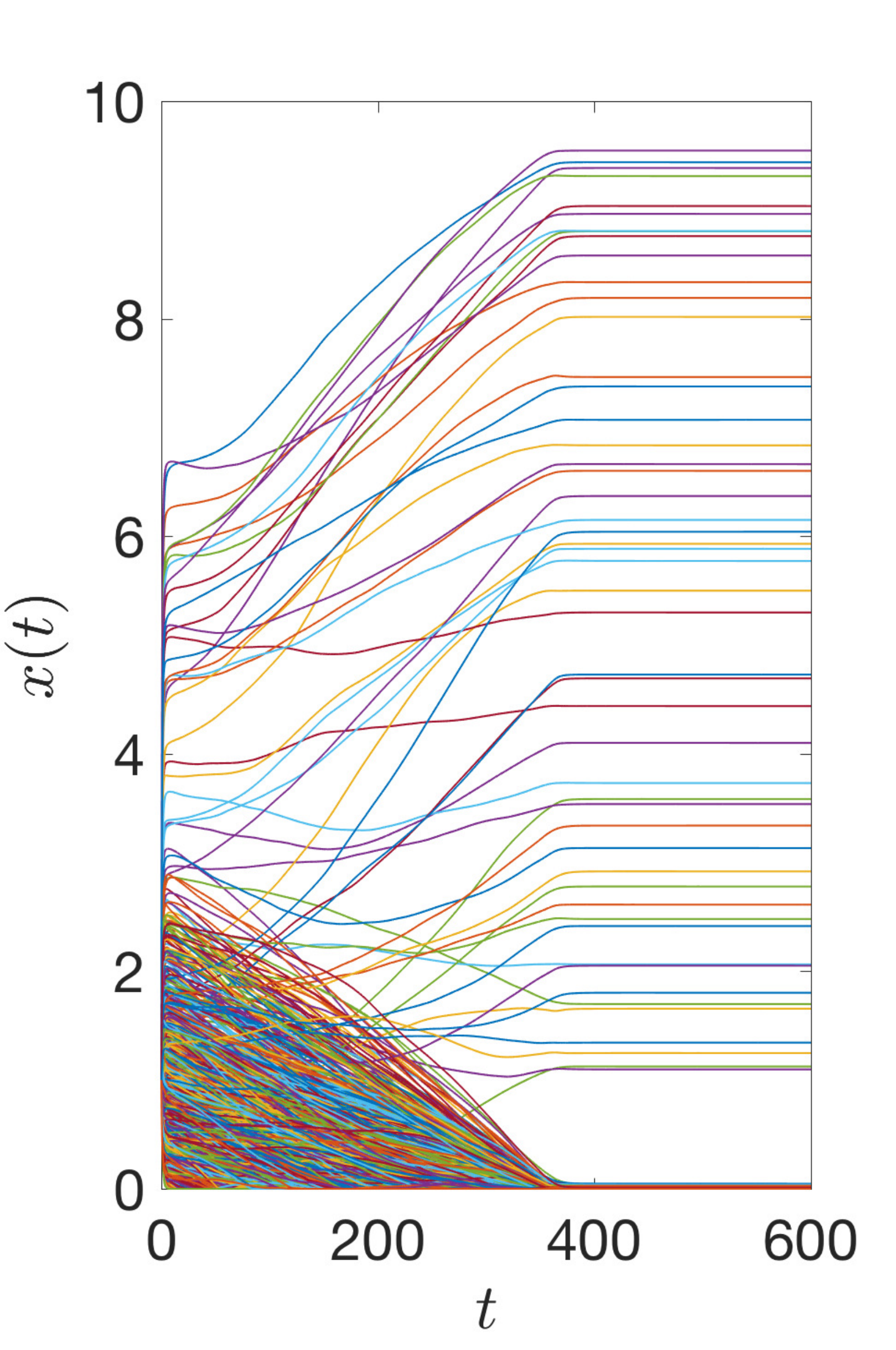}
\label{fig_first_case}}
\hfil
\subfigure[]{\includegraphics[width=1in]{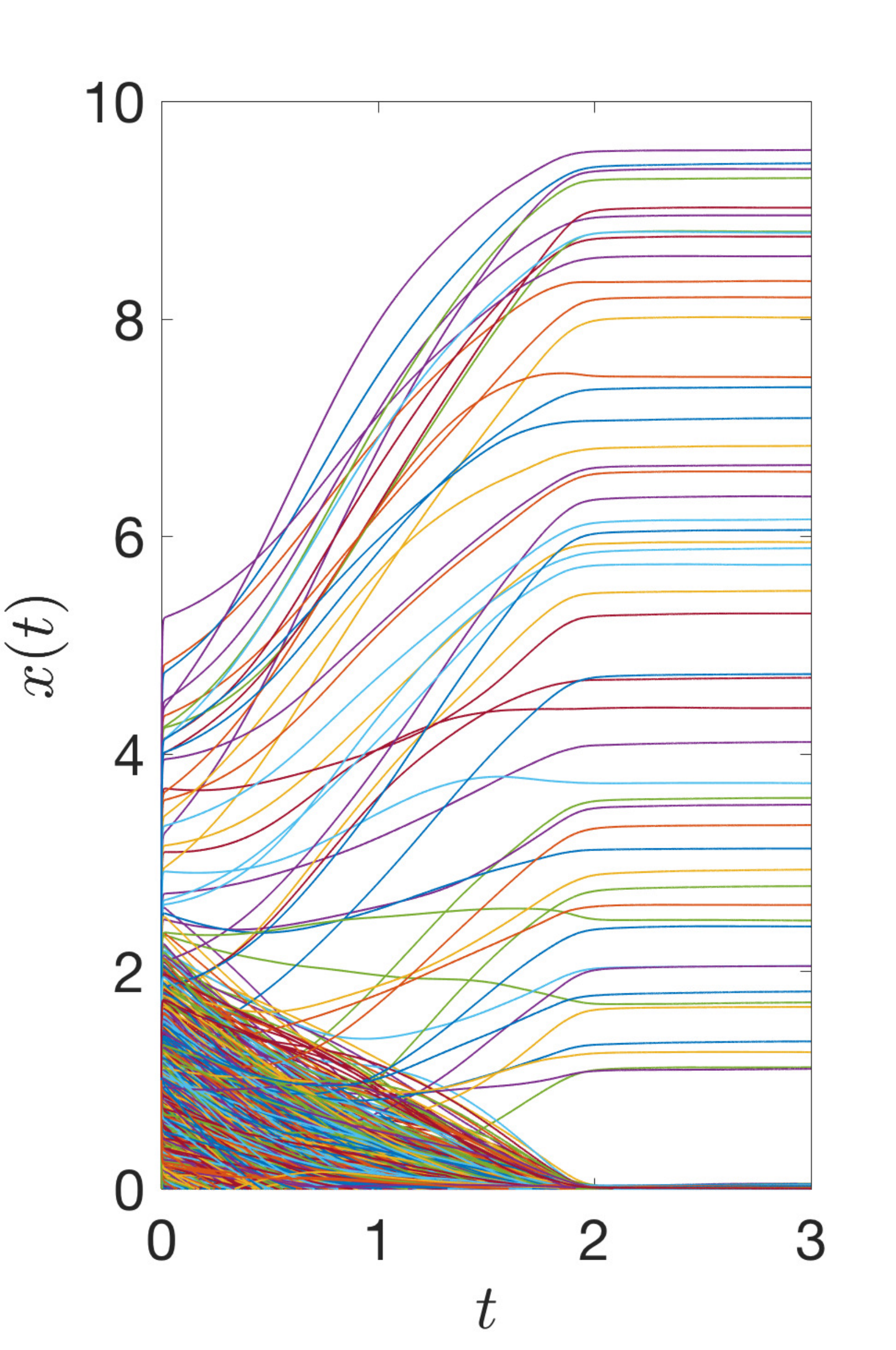}
\label{fig_second_case}}
\hfil
\subfigure[]{\includegraphics[width=1in]{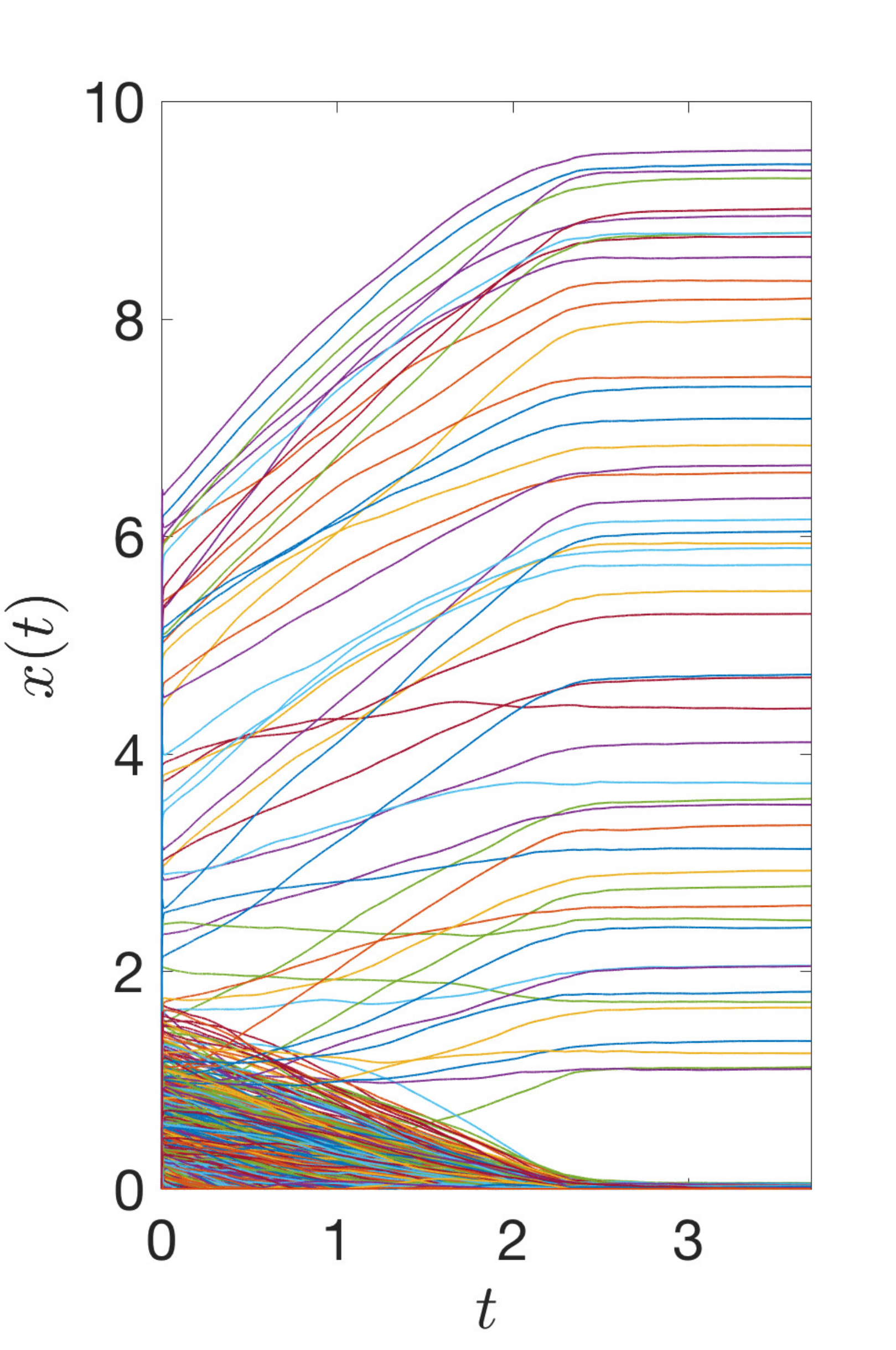}
\label{fig_third_case}}
\caption{State trajectories of (a) neural network (\ref{Neural Network}), (b) SNN and (c) LPNN.} \label{statess}
\end{figure}

\begin{figure}
\centering
\includegraphics[width=2.5in]{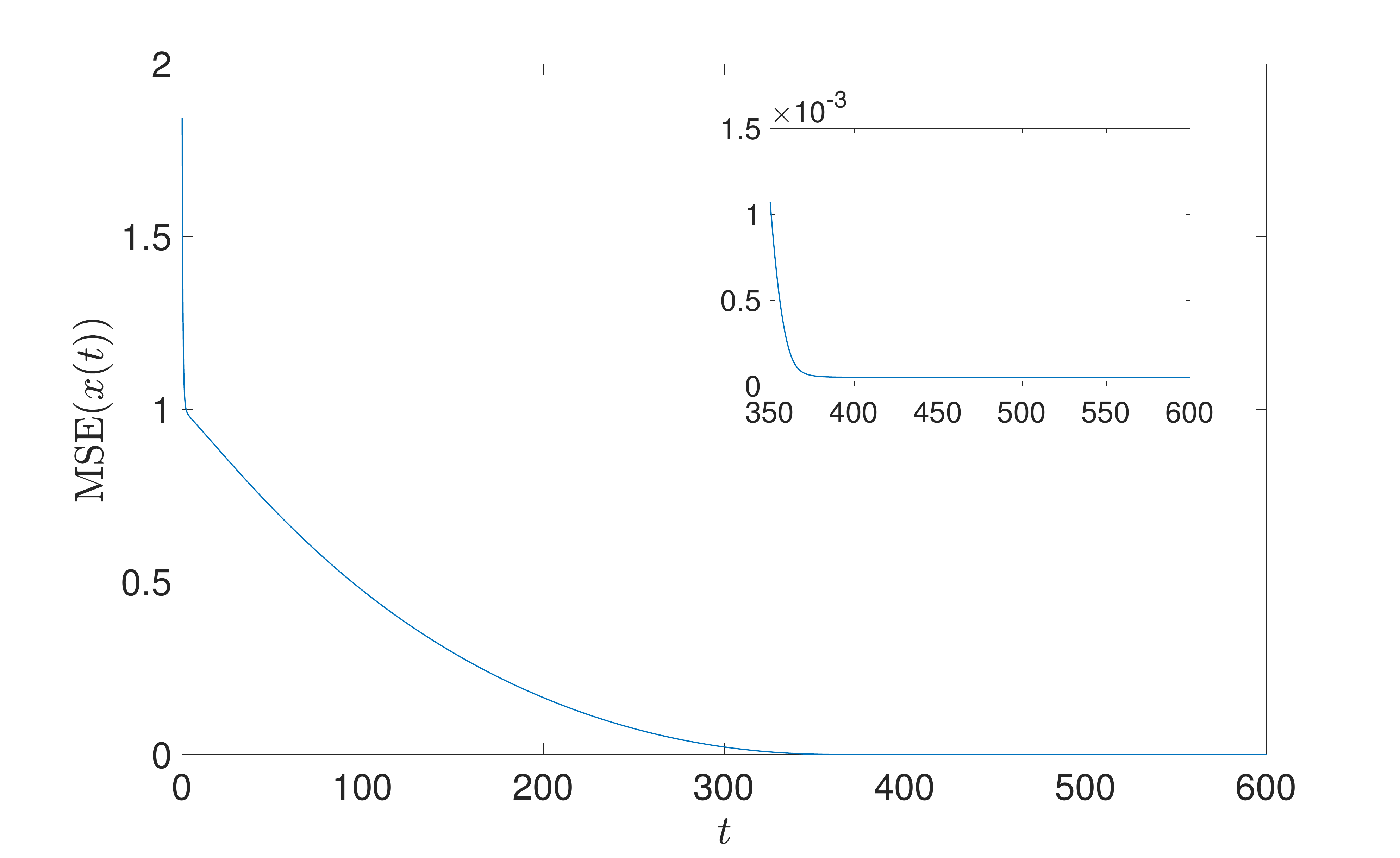}
\caption{Mean squared error of $x(t)$ to the original signal.}
\label{fig:useregressionMSE}
\end{figure}

\begin{figure}
\centering
\includegraphics[width=2.5in]{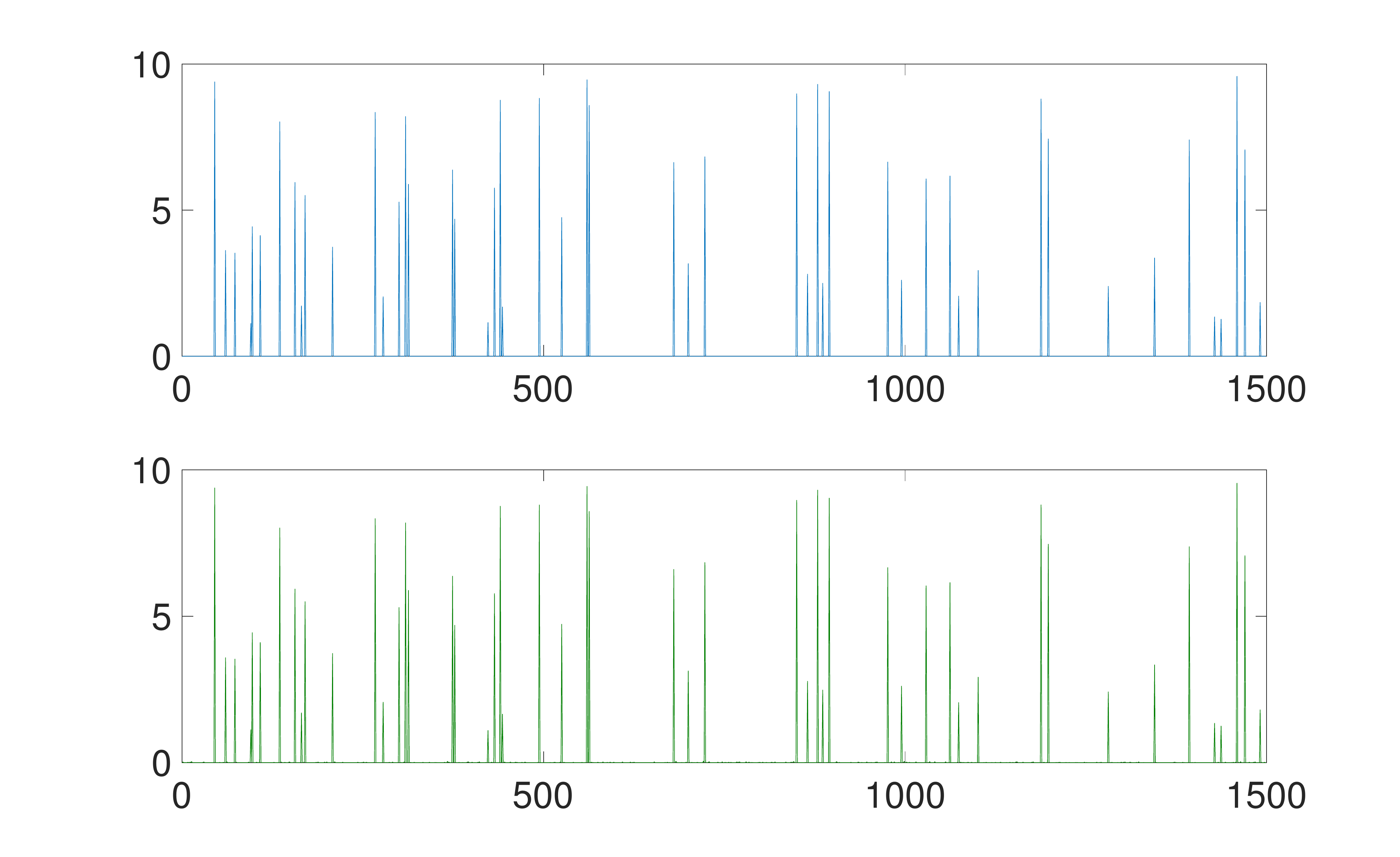}
\caption{Original signal and output solution by neural network \reff{Neural Network} for problem \reff{regression}, where the one above is original signal.}
\label{fig:useregressionplot}
\end{figure}

We compare the proposed neural network (\ref{Neural Network}) with two state-of-the-art neural networks SNN ($p=1.5$) in \cite{Zhao2020} and CBPDN-LPNN (denoted by LPNN in the following) in \cite{Feng2017} with the initial point of all ones. SNN and LPNN are proposed for solving the $L_1$-penalized problems with known noise level, which are continuous convex optimization models. But neural network (\ref{Neural Network}) is used to solve the discontinuous and nonconvex problem with unknown noise level. We run the SNN and LPNN with the true value of noise level in generating the data. The state trajectories of network (\ref{Neural Network}), SNN and LPNN are as shown in Figure \ref{statess}, respectively. It shows that the recovery performance of network (\ref{Neural Network}), SNN and LPNN at the corresponding characteristics time $t$ are stable. Then, we randomly generated ten sets of data for numerical comparison. As can be seen from Table \ref{compare2}, network (\ref{Neural Network}) runs the least CPU time when these neural networks reach good and stable numerical results. Moreover, as shown in Fig. \ref{fig:useregressionMSE}, the mean squared error of $x(t)$ to the original signal is decreasing along the solution of \reff{Neural Network}. The comparison between output solution and original signal can be seen in Fig. \ref{fig:useregressionplot}, which shows that they are almost the same.
\begin{table}
\renewcommand{\arraystretch}{1.3}
\caption{MSEs and CPU time with different neural networks}
\label{compare2}
\centering \begin{tabular}{|c||c|c|c|}
\hline
Network& (\ref{Neural Network})& SNN&LPNN \\
\hline
Mean-MSE & $5.208\times 10^{-5}$ & $8.180\times 10^{-5}$ &$1.130\times 10^{-4}$ \\\hline
Max-MSE & $5.966\times 10^{-5}$ & $9.202\times 10^{-5}$ &$1.415\times 10^{-4}$\\\hline
Min-MSE & $4.747\times 10^{-5}$ & $7.804\times 10^{-5}$ &$9.383\times 10^{-5}$ \\\hline
Mean-CPU(s) & $12.3647$ & $44.4610$&$43.8776$ \\\hline
Max-CPU(s) & $14.1424$ & $49.3730$&$52.7790$ \\\hline
Min-CPU(s)& $11.7356$ & $39.2379$&$38.6471$ \\\hline
\end{tabular}
\end{table}
\subsection{Prostate cancer}
In this experiment, we consider the problem on finding the most important predictors in predicting the prostate cancer. The prostate cancer data set is from the https://web.stan- ford.edu/\textasciitilde hastie/ElemStatLearn/data.html and includes the medical records of 97 men who were plan to receive a radical prostatectomy. This data set is divided into two sets, i.e. a training set with 67 observations and a test set with 30 observations. More detailed explanation on the prostate cancer data set can be found in \citet{Chen-MP}, \citet{Stamey1989Prostate} and \citet{Hastie2017}. The prediction error is defined by the mean squared error over the test set.

In order to solve this problem, we consider the following sparse regression model:
\begin{equation}\label{R}
\begin{split}
&\min~\|Ax-b\|^2+\lambda\|x\|_0\\
&~\text{s.t. }~  x\in\mathcal{X}:=\{x\in\mathbb{R}^{n}:-\upsilon\leq{x}\leq\upsilon\},
\end{split}
\end{equation}
where $A\in\mathbb{R}^{67\times n}$ and $b\in\mathbb{R}^{67}$ are composed by the training set, $n=8$, $\lambda=2$ and $\upsilon=\textbf{10}_{n}$.

We use neural network \reff{Neural Network} to solve the equivalent problem of \reff{R} as in Proposition \ref{xx+x-}. Choose $\alpha_0=20$, $\beta=2$, $\gamma=1$ and initial point $x_0=\textbf{1}_{16}$ in network \reff{Neural Network}. We report the numerical results in Table \ref{result3}, where the listed result for network \reff{Neural Network} is the output solution by \reff{Neural Network} at $t=25$, the result for FOIPA is the best result from Table $2$ in \cite{Bian2015Complexity}, and the results for Lasso and Best subset selection are the best two results from Table $3.3$ in \cite{Hastie2017}. From Table \ref{result3}, we see that the proposed network not only finds the right main predictors in predicting prostate cancer, but also finds a solution with the smallest prediction error among the four methods.
\begin{table}
\renewcommand{\arraystretch}{1.3}
\caption{Results with different methods}
\label{result3}
\centering \begin{tabular}{|c||c|c|c|c|}
\hline
Method& \reff{Neural Network} & FOIPA& Lasso & Best subset\\
\hline
$x^*_1(\rm lcavol)$& 0.6134 & 0.6497 & 0.533 & 0.740 \\
\hline
$x^*_2(\rm lweight)$& 0.3156 & 0.2941 & 0.169 & 0.316 \\
\hline
$x^*_3(\rm age)$& 0 & 0 & 0 & 0 \\
\hline
$x^*_4(\rm lbph)$& 0 & 0 & 0.002 & 0 \\
\hline
$x^*_5(\rm svi)$& 0.2222 & 0.1498 & 0.094 & 0 \\
\hline
$x^*_6(\rm lcp)$& 0 & 0 & 0 & 0 \\
\hline
$x^*_7(\rm gleason)$& 0 & 0 & 0 & 0 \\
\hline
$x^*_8(\rm pgg45)$& 0 & 0 & 0 & 0 \\
\hline
$\|x^*\|_0$& 3 & 3 & 4 & 2 \\
\hline
Prediction error& 0.4002 & 0.4194 & 0.479 & 0.492 \\
\hline
\end{tabular}
\end{table}
\section {Conclusions}
In this paper, we studied a class of sparse regression problem with cardinality penalty. By constructing a smoothing function for cardinality function, we proposed the projection neural network \reff{Neural Network} to solve sparse regression problem \reff{optimization model}. We proved that the solution of \reff{Neural Network} is unique, global existent, bounded and globally Lipschitz continuous. Moreover, we proved that all accumulation points of \reff{Neural Network} have a common support set and own a unified lower bound for the nonzero elements. Furthermore, we proposed a correction method for its accumulation points to obtain the local minimizers of \reff{optimization model}. Specially, in most cases, by using \reff{Neural Network}, a local optimal solution of \reff{optimization model} with lower bound property can be obtained without using the correction. Besides, we proved that the equivalence on local minimizers between \reff{optimization model} and another sparse regression model \reff{two side0}. Finally, some numerical experiments were provided to show the convergence and efficiency of \reff{Neural Network} for solving \reff{optimization model} and \reff{two side0}.
\appendix
\section{Proof of Proposition \ref{prop1}}\label{a1}
\begin{pf}
It is clear that $\theta(\cdot, \mu)$ is a bounded function on $\mathbb{R}^n_+$ and $\lim_{\mu\downarrow0}\theta(s, \mu)=1$ as $s>0$, and $\theta(s, \mu)=0$ for any $\mu>0$ as $s=0$.
Then, $$\lim_{\mu\downarrow0}\Theta(x, \mu)=\sum\limits_{i=1}^{n}\lim_{\mu\downarrow0}\theta(x_i, \mu)=\|x\|_0,\,\,\forall x\in\mathbb{R}^n_+.$$
Since $\theta(s, \mu)$ is differentiable with respect to $s$ for any fixed $\mu>0$, $\Theta(x, \mu)$ is differentiable with respect to $x$ for any fixed $\mu>0$. Thus, $\Theta(x,\mu)$ is a smoothing function of $\|x\|_0$ on $\mathbb{R}^n_+$.
	
Next, we prove the other results in this proposition one by one.
	
As can be seen,
\begin{gather*}\label{Derivative}
\nabla_x\Theta(x, \mu)=\sum_{i=1}^{n}\nabla_x\theta(x_i, \mu)=\left(
{\begin{split}
&\nabla_{x_1}\theta(x_1, \mu)\\
&\nabla_{x_2}\theta(x_2, \mu)\\
&~~~~~~~\vdots\\
&\nabla_{x_n}\theta(x_n, \mu)\\
\end{split}}\right).
\end{gather*}
For any fixed $\mu>0$, since
\begin{gather}\label{Respectively Derivative x2}
\nabla_{x_i}\theta(x_i, \mu)=\left\{
\begin{split}
&\frac{3}{2\mu}~~~~~~\quad \qquad {\rm if}~{x_i}<\frac{1}{3}\mu,\\
&\frac{9}{4\mu^2}(\mu-x_i)\quad {\rm if}~\frac{1}{3}\mu\leq{x_i}\leq\mu,\\
&0~~~~~~~~~~~~~~\qquad {\rm if}~{x_i}>\mu,
\end{split}
\right.
\end{gather}we have $\nabla_{x_i}\theta(\cdot, \mu)$ is continuous for any $ i=1,2,,n$. Then, $\Theta(\cdot, \mu)$ is continuously differentiable for any fixed $\mu>0$.
Moreover, since
\begin{gather*}
\nabla_{x_i}^2\theta(x_i, \mu)=\left\{
\begin{split}
&0~~\qquad \qquad {\rm if}~{x_i}<\frac{1}{3}\mu,\\
&-\frac{9}{4\mu^2}~\qquad {\rm if}~\frac{1}{3}\mu<{x_i}<\mu,\\
&0~~~~~~~~~~~\qquad {\rm if}~{x_i}>\mu,
\end{split}
\right.
\end{gather*}
we obtain that for any fixed $\mu>0$, $\nabla_{x_i}^2\theta(\cdot, \mu)$ is bounded on $\left(-\infty,\frac{1}{3}\mu\right)\cup\left(\frac{1}{3}\mu,\mu\right)\cup\left(\mu,+\infty \right)$. Then, $\nabla_{x_i}\theta(\cdot, \mu)$ is globally Lipschitz continuous on $\mathbb{R}$ for any $i=1,2,\ldots,n$. Therefore $\nabla_x\Theta(\cdot, \mu)$ is globally Lipschitz continuous on $\mathbb{R}^n$ for any fixed $\mu>0$, which means that result (i) in this proposition holds.
	
For any fixed $x\in\mathbb{R}_+^n$, we also see that
\begin{equation*}
\nabla_{\mu}\Theta(x,\mu)=\sum\limits_{i=1}^{n}\nabla_{\mu}\theta(x_i,\mu),
\end{equation*}
with
\begin{gather}\label{Respectively Derivative mu2}
\nabla_{\mu}\theta(x_i, \mu)=\left\{
\begin{split}
&-\frac{3x_i}{2\mu^2}~~\quad \qquad {\rm if}~\mu>3 x_i,\\
&-\frac{9(\mu-x_i)x_i}{4\mu^3}~{\rm if}~x_i\leq\mu\leq3x_i,\\
&0~~~~~~~~~~~~~~~\qquad {\rm if}~\mu<x_i.
\end{split}
\right.
\end{gather}Then, $\nabla_{\mu}\Theta(x, \cdot)$ is continuous on $(0,+\infty)$, which implies $\Theta(x, \cdot)$ is continuously differentiable on $(0,+\infty)$ for any fixed $x\in\mathbb{R}^n_+$.
Moreover, since
\begin{gather*}
\nabla_{\mu}^2\theta(x_i, \mu)=\left\{
\begin{split}
&\frac{3x_i}{\mu^3}~~~~~~\qquad \qquad {\rm if}~\mu>3x_i,\\
&\frac{9x_i(2\mu-3x_i)}{4\mu^4}~~~~ {\rm if}~x_i<\mu<3x_i,\\
&0~~~~~~~~~~~~~~~~~\qquad {\rm if}~\mu<x_i,
\end{split}
\right.
\end{gather*}
which means that $\nabla_{\mu}\Theta(x, \cdot)$ is locally Lipschitz continuous on $(0,+\infty)$. Thus, property (ii) holds.
	
Since $\mathcal{X}$ and $[\underline{\mu},\bar{\mu}]$ are compact sets in $\mathbb{R}_+^n$ and $\mathbb{R}_+$, respectively, by \reff{Respectively Derivative x2} and \reff{Respectively Derivative mu2}, we obtain result (iii) in this proposition.
\end{pf}
\section{Proof of Lemma \ref{lemma1}}\label{a2}
\begin{pf}
Let $w =x_t-\nabla f(x_t)-\lambda\nabla_x\Theta\left(x_{t}, \mu_t\right)$ and $u=x_t$ in Proposition \ref{projection operator property}, by \reff{Neural Network}, then we obtain
\begin{align}\label{xt}
\left\langle\nabla{f}\left(x_{t}\right)+\lambda\nabla_x\Theta\left(x_{t}, \mu_t\right),\dot{x}_{t}\right\rangle\leq-\frac{1}{\gamma}\|\dot{x}_{t}\|^{2}.
\end{align}
Since $\mu_t\geq\frac{1}{2}\mu^*$ and $\dot{\mu}_t\leq0$, by result (iii) of Proposition \ref{prop1}, there exists $\varrho>0$ such that
\begin{align}\label{ut}
\nabla_\mu\Theta\left(x_{t}, \mu_t\right){\mu}_{t}'\leq-\varrho{\mu}_{t}'.
\end{align}
Since \begin{small}
\begin{equation}\label{3}
\begin{split}
&\frac{d}{dt}\left[f(x_{t})+\lambda\Theta\left(x_{t}, \mu_t\right)\right]\\
=~&\left\langle\nabla{f}\left(x_{t}\right)+\lambda\nabla_x\Theta\left(x_{t}, \mu_t\right),\dot{x}_{t}\right\rangle+\lambda\left\langle\nabla_\mu\Theta\left(x_{t}, \mu_t\right),{\mu}_{t}'\right\rangle,
\end{split}
\end{equation}
\end{small}by \reff{xt} and \reff{ut}, we obtain
\begin{small}
\begin{align}\label{non-increase}
\frac{d}{dt}\left[f(x_{t})+\lambda\Theta\left(x_{t}, \mu_t\right)+\lambda\varrho\mu_t\right]\leq-\frac{1}{\gamma}\|\dot{x}_{t}\|^{2}.
\end{align}
\end{small}Thus, $f(x_{t})+\lambda\Theta\left(x_{t}, \mu_t\right)+\lambda\varrho\mu_t$ is nonincreasing on $[0,+\infty)$. Since $f(x)$ and $\Theta\left(x, \mu\right)$ are continuous on $\mathcal{X}\times[\frac{1}{2}\mu^*,\mu_0]$, $x_t\in \mathcal{X}$ and $\mu_t\in[\frac{1}{2}\mu^*,\mu_0]$, we get that $f(x_{t})+\lambda\Theta\left(x_{t}, \mu_t\right)+\lambda\varrho\mu_t$ is bounded from below on $[0,+\infty)$. As a consequence, $\lim_{t\rightarrow+\infty}\left[f(x_{t})+\lambda\Theta\left(x_{t}, \mu_t\right)+\lambda\varrho\mu_t\right]$ exists. Taking into account that $\lim_{t\rightarrow+\infty}\varrho\mu_t=\frac{\varrho\mu^*}{2}$, we conclude that \begin{equation*}
\lim_{t\rightarrow+\infty}\left[f(x_{t})+\lambda\Theta\left(x_{t}, \mu_t\right)\right]
\end{equation*} exists.
Using (\ref{non-increase}) again, we have $\int_0^{+\infty}\|\dot{x}_t\|^2 dt<+\infty$.
	
Recalling the existence of $\lim_{t\rightarrow+\infty}[f(x_{t})+\lambda\Theta\left(x_{t}, \mu_t\right)+\lambda\varrho\mu_t]$ again, by \reff{non-increase}, we have that
\begin{equation}\label{4}
\int_0^{+\infty}\left|\frac{d}{dt}\left[f(x_{t})+\lambda\Theta\left(x_{t}, \mu_t\right)+\lambda\varrho\mu_t\right]\right|dt<+\infty.
\end{equation}
	
Similar to the proof in Theorem \ref{theorem1}, $\nabla_{\mu}\Theta\left(x_{t}, \mu_t\right)$ is globally Lipschitz continuous on $[0,+\infty)$. Combining this with global Lipschitz continuity of $\dot{x}_t$, $\dot{\mu}_t$, $\nabla{f}\left(x_{t}\right)$ and $\nabla_x\Theta\left(x_{t}, \mu_t\right)$ on $[0,+\infty)$, by \reff{3}, we deduce $\frac{d}{dt}\left[f(x_{t})+\lambda\Theta\left(x_{t},\mu_t\right)+\lambda\varrho\mu_t\right]$ is globally Lipschitz continuous on $[0,+\infty)$, which is of course uniform continuous on $[0,+\infty)$. Recalling \reff{4} and Proposition \ref{limit}, we conclude \begin{equation*}
\lim_{t\rightarrow+\infty}\frac{d}{dt}[f(x_{t})+\lambda\Theta\left(x_{t}, \mu_t\right)+\varrho\mu_t]=\bm{0}.
\end{equation*}
Returning to inequality \reff{non-increase}, we deduce that \begin{equation*}
\lim_{t\rightarrow+\infty}\dot{x}_t=\bm{0}.
\end{equation*}
\end{pf}
\section{Proof of Lemma \ref{lemma2}}\label{a3}
\begin{pf}
From the definition of $\mu_t$, we have that there exists a $T_\mu>0$ such that $\mu_t\in\left(\frac{1}{2}\mu^*,\mu^*\right)$ for any $t> T_\mu$. 
	
We prove the first result by contradiction. Suppose that some entry of $x_t$ denoted by $[x_t]_{i_0}$ is less than or equal to $\frac{1}{6}\mu^*$ at some point $T\geq T_\mu$, i.e. $[x_T]_{i_0}\leq\frac{1}{6}\mu^*$, which implies $[x_T]_{i_0}<\frac{1}{3}\mu_{T}$. Returning to \reff{Respectively Derivative x2}, we have $\nabla_{x_{i_0}}\theta([x_T]_{i_0},\mu_T)=\frac{3}{2\mu_T}$. Further, based on ${\bar{v}+L_f}-{\lambda}\frac{3}{2\mu^*}<0$ by Assumption \ref{choice}, we have $\left[x_T-\nabla{f}\left(x_T\right)-\lambda\nabla_x\Theta\left(x_T,\mu_T\right)\right]_{i_0}<0$,
which means, $\left[P_{\mathcal{X}}\left[x_T-\nabla{f}\left(x_T\right)-\lambda\nabla_x\Theta\left(x_T,\mu_T\right)\right]\right]_{i_0}=0$,
and hence
\begin{align}\label{x=-}
[\dot x_T]_{i_0}=-\gamma [x_T]_{i_0}.
\end{align}
Since \reff{x=-} holds for any $[x_T]_{i_0}$ satisfying $[x_T]_{i_0}\leq\frac{1}{6}\mu^*$, from the non-increasing property of it deduced by \reff{x=-}, we obtain $[\dot x_t]_{i_0}=-\gamma [x_t]_{i_0},\forall t\geq T$. Therefore for any ${t}>T$, ${[x_t]_{i_0}}={[x_T]_{i_0}}e^{-\gamma(t-T)}$,
which tends to $0$ as $t$ tends to $+\infty$. Therefore, if some entry of $x_t$ is less than or equal to $\frac{1}{6}\mu^*$ at some point $T\geq T_\mu$, then the limit of this entry is 0.
	
Conversely, if some entry of $x_t$ is more than $\frac{1}{6}\mu^*$ for any $t\geq T_{\mu}$, then any accumulation point of this entry is not less than $\frac{1}{6}\mu^*$. 
	
As a result, for any accumulation point $\bar{x}$ of $x_{t}$, either $\bar{x}_i=0$ or $\bar{x}_i\geq\frac{1}{6}\mu^*$, $i=1,\ldots,n$, and hence $I(\bar{x})=\{i\in\{1,2,\ldots,n\}:\bar{x}_i=0\}$.
	
In addition, if there exists an $\hat{i}\in\{1,2,\ldots,n\}$ such that $0$ is an accumulation point of $[x_t]_{\hat{i}}$, then there exists some point $\hat{T}\geq T_{\mu}$ such that	$[x_{\hat{T}}]_{\hat{i}}\leq\frac{1}{6}\mu^*$, which implies $0$ is the unique accumulation point of $[x_t]_{\hat{i}}$. Therefore, for any accumulation points $\hat{x}$ and $\tilde{x}$ of $x_{t}$, we have $I(\hat{x})=I(\tilde{x})$.
\end{pf}
\section{Proof of Proposition \ref{==}}\label{a4}
\begin{pf}
If $I(x^*)=\phi$, then the equivalence is obviously true. Next, we consider the case $I(x^*)\neq\phi$. Let $\sigma=\min\{x^*_i:x^*_i\neq0,i=1,2,...,n\}$ and $\delta=\frac{\sigma}{2}$, then 
\begin{align}\label{*=}
||x||_0=||x^*||_0,\forall x\in \mathcal{X}_{I}(x^*)\cap B_\delta(x^*).
\end{align}
If $x^*$ is a local minimizer of function $f(x)+\lambda||x||_0$ in $\mathcal{X}$, then there is a $\delta_1\in(0,\delta]$ such that 
\begin{align}\label{**=}
f(x)+\lambda||x||_0\geq f(x^*)+\lambda||x^*||_0,\forall x\in \mathcal{X}\cap B_{\delta_1}(x^*).
\end{align}
Let $\delta_2\in(0,\delta_1]$, by \reff{*=} and \reff{**=}, then we obtain \begin{equation*}
f(x)\geq f(x^*),\forall x\in \mathcal{X}_{I}(x^*)\cap B_{\delta_2}(x^*).
\end{equation*}
Therefore $x^*$ is a local minimizer of $f(x)$ in $\mathcal{X}_{I}(x^*)$.
	
Conversely, if $x^*$ is a local minimizer of $f(x)$ in $\mathcal{X}_{I}(x^*)$, then there exists a $\delta_3\in(0,\delta]$ such that
\begin{align}\label{>=}
f(x)\geq f(x^*),\forall x\in \mathcal{X}_{I}(x^*)\cap B_{\delta_3}(x^*). 
\end{align}Since $f$ is continuous in $\mathcal{X}$, for any $\varepsilon\in(0,\lambda)$, there exists a $\delta_4\in(0,\delta_3]$ such that $|f(x)-f(x^*)|<\varepsilon,\forall x\in\mathcal{X}\cap B_{\delta_4}(x^*)$, which implies 
\begin{align}\label{<<}
f(x^*)<f(x)+\varepsilon<f(x)+\lambda,\forall x\in\mathcal{X}\cap B_{\delta_4}(x^*).
\end{align}
Combining \reff{*=} with \reff{>=}, we get $$f(x)+\lambda||x||_0\geq f(x^*)+\lambda||x^*||_0,\forall x\in \mathcal{X}_{I}(x^*)\cap B_{\delta_4}(x^*).$$ For any $x\in \mathcal{X}\cap B_{\delta_4}(x^*)$ and $x\notin\mathcal{X}_{I}(x^*)$, we have \begin{equation*}
||x||_0\geq||x^*||_0+1
\end{equation*}and hence by \reff{<<}, we obtain
\begin{align*}
f(x)+\lambda||x||_0&\geq f(x)+\lambda(||x^*||_0+1)\\&=f(x)+\lambda+\lambda||x^*||_0\\&>f(x^*)+\lambda||x^*||_0.
\end{align*}
Therefore $x^*$ is a global minimizer of $f(x)+\lambda||x||_0$ in $\mathcal{X}\cap B_{\delta_4}(x^*)$. As a result, $x^*$ is a local minimizer of $f(x)+\lambda||x||_0$ in $\mathcal{X}$.
\end{pf}
\section{Proof of Proposition \ref{accumulation}}\label{a5}
\begin{pf}
From the continuity of $f$, there exists a $\delta>0$ such that
\begin{align}\label{<<<}
f(\bar{x}^{\mu^*})<f(x)+\lambda,\forall x\in B_{\delta}(\bar{x}^{\mu^*}).
\end{align}
For any $x\in \mathcal{X}_{K}(\bar{x})$ and $x\neq\bar{x}^{\mu^*}$, we have
\begin{align}\label{+1}
\|x\|_0\geq\|\bar{x}^{\mu^*}\|_0+1.
\end{align}
Combining \reff{<<<} and \reff{+1}, for any $x\in B_{\delta}(\bar{x}^{\mu^*})\cap \mathcal{X}_{K}(\bar{x})$ and $x\neq\bar{x}^{\mu^*}$, we deduce
\begin{align*}
\begin{split}
f(x)+\lambda\|x\|_0&\geq{f(x)}+\lambda(\|\bar{x}^{\mu^*}\|_0+1)\\&=(f(x)+\lambda)+\lambda\|\bar{x}^{\mu^*}\|_0\\&>f(\bar x^{\mu^*})+\lambda\|\bar{x}^{\mu^*}\|_0,
\end{split}
\end{align*}
which means that $\bar{x}^{\mu^*}$ is a strictly local minimizer of $f(x)+\lambda\|x\|_0$ in $\mathcal{X}_{K}(\bar{x})$.
	
If $J(\bar{x})=\emptyset$, then $\bar{x}=\bar{x}^{\mu^*}$ and hence $f(\bar{x}^{\mu^*})+\lambda\|\bar{x}^{\mu^*}\|_0= f(\bar{x})+\lambda\|\bar{x}\|_0$. Next, we consider the case of $J(\bar{x})\neq\emptyset$. 
	
In view of $\sup_{x\in{\mathcal{X}}}\|\nabla{f(x)}\|_\infty\leq L_f$, we can regard $\sqrt{n}L_f$ as a Lipschitz constant of $f$ on $\mathcal{X}$, which means \begin{align}\label{||<}
|f(x)-f(\bar{x})|\leq \sqrt{n}L_f\|x-\bar{x}\|,\forall x\in \mathcal{X}.
\end{align}
If $\|x-\bar{x}\|<\lambda/(\sqrt{n}{L_f})$, then by \reff{||<}, we have $|f(x)-f(\bar{x})|<\lambda$. Let $\bar{\delta}=\lambda/(\sqrt{n}{L_f})$, then we have $|f(x)-f(\bar{x})|<\lambda, \forall x\in B_{\bar{\delta}}(\bar{x})$.
It follows from Assumption \ref{choice}, i.e. ${\sqrt{n}\mu^*}/{2}<\lambda/(\sqrt{n}{L_f})$, that $\bar{x}^{\mu^*}\in B_{\bar{\delta}}(\bar{x})$, which implies
\begin{align}\label{bar*}
f(\bar{x}^{\mu^*})-f(\bar{x})<\lambda.
\end{align}Furthermore, if $J(\bar{x})\neq\emptyset$, we have \begin{align}\label{>1}
\|\bar{x}\|_0-\|\bar{x}^{\mu^*}\|_0\geq1.
\end{align}Combining \reff{bar*} with \reff{>1}, we obtain \begin{equation*}
f(\bar{x}^{\mu^*})-f(\bar{x})<\lambda\left(\|\bar{x}\|_0-\|\bar{x}^{\mu^*}\|_0\right),
\end{equation*} and hence $f(\bar{x}^{\mu^*})+\lambda\|\bar{x}^{\mu^*}\|_0<f(\bar{x})+\lambda\|\bar{x}\|_0$.
\end{pf}
\section{Proof of Proposition \ref{xx+x-}}\label{a6}
\begin{pf}
Let $y^*$ be a local minimizer of $f(y)+\lambda\|y\|_0$ in $\mathcal{Y}$, then there exists a $\delta>0$ such that $f(y^*)+\lambda\|y^*\|_0\leq f(y)+\lambda\|y\|_0$ for any $y\in B_\delta(y^*)\cap \mathcal{Y}$.
	
Since there exist unique $x^*_+\in\mathcal{X}_1$ and $x^*_-\in \mathcal{X}_2$ such that $y^*=x^*_+-x^*_-$ and $(x^*_+)_i(x^*_-)_i=0,i=1,2,\ldots,n$, we obtain that
\begin{align}\label{0=}
\|y^*\|_0=\|x^*_+\|_0+\|x^*_-\|_0.
\end{align}For any $x_+\in B_{\frac{\delta}{2}}(x^*_+)\cap \mathcal{X}_1$ and $x_-\in B_{\frac{\delta}{2}}(x^*_-)\cap \mathcal{X}_2$, we have $x_+-x_-\in \left(B_{\frac{\delta}{2}}(x^*_+)\cap \mathcal{X}_1\right)-\left(B_{\frac{\delta}{2}}(x^*_-)\cap\mathcal{X}_2\right)\subseteq B_\delta(x^*)\cap \mathcal{Y}$. Therefore, for any $x_+\in B_{\frac{\delta}{2}}(x^*_+)\cap \mathcal{X}_1$ and $x_-\in B_{\frac{\delta}{2}}(x^*_-)\cap\mathcal{X}_2$, we have \begin{align}\label{f<=}
f(y^*)+\lambda\|y^*\|_0\leq f(x_+-x_-)+\lambda\|x_+-x_-\|_0.
\end{align}If there exists an $i_0\in\{1,2,...,n\}$ such that $[x_+-x_-]_{i_0}\neq0$, then at least one of $[x_+]_{i_0}$ and $[x_-]_{i_0}$ is not 0. Thus
\begin{align}\label{<=0}
\|x_+-x_-\|_0\leq\|x_+\|_0+\|x_-\|_0.
\end{align}
Using \reff{0=}, \reff{f<=} and \reff{<=0}, we conclude that for any $x_+\in B_{\frac{\delta}{2}}(x^*_+)\cap \mathcal{X}_1$ and $x_-\in B_{\frac{\delta}{2}}(x^*_-)\cap\mathcal{X}_2$,
\begin{align*}
\begin{split}
&f(x^*_+-x^*_-)+\lambda\|x^*_+\|_0+\lambda\|x^*_-\|_0\\=~&{f(x^*)}+\lambda\|{x}^{*}\|_0\\\leq~&f(x_+-x_-)+\lambda\|{x}_+-x_-\|_0\\\leq~&f(x_+-x_-)+\lambda\|{x}_+\|_0+\lambda\|{x}_-\|_0.
\end{split}
\end{align*}
Therefore, $({x^*_+}^\top,{x^*_-}^\top)^\top$ is a local minimizer of $f(x_+-x_-)+\lambda\|x_+\|_0+\lambda\|x_-\|_0$ in $\mathcal{X}_1\times\mathcal{X}_2$.
	
Conversely, assume $({x^*_+}^\top,{x^*_-}^\top)^\top$ is a local minimizer of $f(x_+-x_-)+\lambda\|x_+\|_0+\lambda\|x_-\|_0$ in $\mathcal{X}_1\times\mathcal{X}_2$. Let \begin{equation*}
I^*=\{i\in\{1,2,\ldots,2n\}: ({x^*_+}^\top,{x^*_-}^\top)^\top_i=0\}
\end{equation*} and \begin{equation*}
J^*=\{i\in\{1,2,\ldots,n\}: (x^*_+)_i=(x^*_-)_i\}.
\end{equation*} Define \begin{equation*}
\left({\mathcal{X}_1\times\mathcal{X}_2}\right)_{I^*}=\{(x^\top_+,x^\top_-)^\top\in \mathcal{X}_1\times\mathcal{X}_2:x_i=0,\forall i\in I^*\}
\end{equation*} and \begin{equation*}
\mathcal{Y}_{J^*}=\{y\in \mathcal{Y}:y_i=0,\forall i\in J^*\}.
\end{equation*}
In view of Proposition \ref{==}, we know that there exists a $\delta'>0$ such that for any $({x_+}^\top,{x_-}^\top)^\top\in B_{\delta'}(({x^*_+}^\top,{x^*_-}^\top)^\top)\cap \left({\mathcal{X}_1\times\mathcal{X}_2}\right)_{I^*}$, $$f(x^*_+-x^*_-)\leq f(x_+-x_-).$$ Let $y^*=x^*_+-x^*_-$. For $y\in B_{\delta'}(x^*)\cap \mathcal{Y}_{J^*}$, there exists $({y_+}^\top,{y_-}^\top)^\top\in B_{\delta'}(({x^*_+}^\top,{x^*_-}^\top)^\top)\cap \left({\mathcal{X}_1\times\mathcal{X}_2}\right)_{I^*}$  such that $y=y_+-y_-$, and hence \begin{equation*}
f(y)=f(y_+-y_-)\geq f(x^*_+-x^*_-)= f(y^*).
\end{equation*}Therefore, $y^*$ is a local minimizer of $f(y)$ in $\mathcal{Y}_{J^*}.$ In view of Proposition \ref{==}, it holds that $y^*$ is a local minimizer of $f(y)+\lambda\|y\|_0$ in $\mathcal{Y}$.
\end{pf}

\bibliographystyle{elsarticle-num-names}

\bibliography{myrefs}

\begin{thebibliography}{59}
\providecommand{\natexlab}[1]{#1}
\providecommand{\url}[1]{\texttt{#1}}
\providecommand{\urlprefix}{URL }
\expandafter\ifx\csname urlstyle\endcsname\relax
  \providecommand{\doi}[1]{doi:\discretionary{}{}{}#1}\else
  \providecommand{\doi}[1]{doi:\discretionary{}{}{}\begingroup
  \urlstyle{rm}\url{#1}\endgroup}\fi
\providecommand{\bibinfo}[2]{#2}

\bibitem[{Candes et~al.(2006)Candes, Romberg, and Tao}]{Candes2006}
\bibinfo{author}{E.~Candes}, \bibinfo{author}{J.~Romberg},
  \bibinfo{author}{T.~Tao}, \bibinfo{title}{Robust uncertainty principles:
  exact signal reconstruction from highly incomplete frequency information},
  \bibinfo{journal}{IEEE Trans. Inf. Theory}
  \bibinfo{volume}{52}~(\bibinfo{number}{1}) (\bibinfo{year}{2006})
  \bibinfo{pages}{489--509}.

\bibitem[{B\"uhlmann et~al.(2014)B\"uhlmann, Kalisch, and Meier}]{B2014High}
\bibinfo{author}{P.~B\"uhlmann}, \bibinfo{author}{M.~Kalisch},
  \bibinfo{author}{L.~Meier}, \bibinfo{title}{High-dimensional statistics with
  a view toward applications in biology}, \bibinfo{journal}{Ann. Rev. Stat.
  Appl.} \bibinfo{volume}{1}~(\bibinfo{number}{1}) (\bibinfo{year}{2014})
  \bibinfo{pages}{255--278}.

\bibitem[{Liu and Wu(2007)}]{Liu2007Variable}
\bibinfo{author}{Y.~Liu}, \bibinfo{author}{Y.~Wu}, \bibinfo{title}{Variable
  selection via a combination of the ${L}_0$ and ${L}_1$ penalties},
  \bibinfo{journal}{J. Comput. Graph. Statist.}
  \bibinfo{volume}{16}~(\bibinfo{number}{4}) (\bibinfo{year}{2007})
  \bibinfo{pages}{782--798}.

\bibitem[{Soubies et~al.(2015)Soubies, Blanc-F\'eraud, and
  Aubert}]{Soubies2015A}
\bibinfo{author}{E.~Soubies}, \bibinfo{author}{L.~Blanc-F\'eraud},
  \bibinfo{author}{G.~Aubert}, \bibinfo{title}{A continuous exact $\ell_0$
  penalty ({CEL}0) for least squares regularized problem},
  \bibinfo{journal}{SIAM J. Imaging Sci.}
  \bibinfo{volume}{8}~(\bibinfo{number}{3}) (\bibinfo{year}{2015})
  \bibinfo{pages}{1607--1639}.

\bibitem[{Hastie et~al.(2017)Hastie, Tibshirani, and Friedman}]{Hastie2017}
\bibinfo{author}{T.~Hastie}, \bibinfo{author}{R.~Tibshirani},
  \bibinfo{author}{J.~Friedman}, \bibinfo{title}{The Elements of Statistical
  Learning Data Mining, Inference, and Prediction, Second Edition},
  \bibinfo{publisher}{New York: Springer-Verlag}, \bibinfo{year}{2017}.

\bibitem[{Thi et~al.(2015)Thi, Dinh, Le, and Vo}]{H2015DC}
\bibinfo{author}{H.~L. Thi}, \bibinfo{author}{T.~P. Dinh},
  \bibinfo{author}{H.~Le}, \bibinfo{author}{X.~Vo}, \bibinfo{title}{{DC}
  approximation approaches for sparse optimization}, \bibinfo{journal}{Eur. J.
  Oper. Res.} \bibinfo{volume}{244}~(\bibinfo{number}{1})
  (\bibinfo{year}{2015}) \bibinfo{pages}{26--46}.

\bibitem[{Nikolova(2016)}]{Nikolova2016Relationship}
\bibinfo{author}{M.~Nikolova}, \bibinfo{title}{Relationship between the optimal
  solutions of least squares regularized with ${L}_0$-norm and constrained by
  k-sparsity}, \bibinfo{journal}{Appl. Comput. Harmon. Anal.}
  \bibinfo{volume}{41}~(\bibinfo{number}{1}) (\bibinfo{year}{2016})
  \bibinfo{pages}{237--265}.

\bibitem[{Natarajan(1995)}]{Natarajan1995Sparse}
\bibinfo{author}{B.~K. Natarajan}, \bibinfo{title}{Sparse approximate solutions
  to linear systems}, \bibinfo{journal}{SIAM J. Comput.}
  \bibinfo{volume}{24}~(\bibinfo{number}{2}) (\bibinfo{year}{1995})
  \bibinfo{pages}{227--234}.

\bibitem[{Chen et~al.(2014)Chen, Ge, Wang, and Ye}]{Chen2014Complexity}
\bibinfo{author}{X.~Chen}, \bibinfo{author}{D.~Ge}, \bibinfo{author}{Z.~Wang},
  \bibinfo{author}{Y.~Ye}, \bibinfo{title}{Complexity of unconstrained
  ${L}_2$-${L}_p$ minimization}, \bibinfo{journal}{Math. Program.}
  \bibinfo{volume}{143}~(\bibinfo{number}{1-2}) (\bibinfo{year}{2014})
  \bibinfo{pages}{371--383}.

\bibitem[{Bian et~al.(2015)Bian, Chen, and Ye}]{Bian2015Complexity}
\bibinfo{author}{W.~Bian}, \bibinfo{author}{X.~Chen}, \bibinfo{author}{Y.~Ye},
  \bibinfo{title}{Complexity analysis of interior point algorithms for
  non-{L}ipschitz and nonconvex minimization}, \bibinfo{journal}{Math.
  Program.} \bibinfo{volume}{149}~(\bibinfo{number}{1-2})
  (\bibinfo{year}{2015}) \bibinfo{pages}{301--327}.

\bibitem[{Liu and Wang(2016)}]{Liu2016}
\bibinfo{author}{Q.~Liu}, \bibinfo{author}{J.~Wang},
  \bibinfo{title}{${L}_1$-minimization algorithms for sparse signal
  reconstruction based on a projection neural network}, \bibinfo{journal}{IEEE
  Trans. Neural Netw. Learn. Syst.} \bibinfo{volume}{27}~(\bibinfo{number}{3})
  (\bibinfo{year}{2016}) \bibinfo{pages}{698--707}.

\bibitem[{Mohimani et~al.(2009)Mohimani, Babaie-Zadeh, and
  Jutten}]{Jutten2008A}
\bibinfo{author}{H.~Mohimani}, \bibinfo{author}{M.~Babaie-Zadeh},
  \bibinfo{author}{C.~Jutten}, \bibinfo{title}{A fast approach for overcomplete
  sparse decomposition based on smoothed $\ell_0$ norm}, \bibinfo{journal}{IEEE
  Trans. Signal Process.} \bibinfo{volume}{57}~(\bibinfo{number}{1})
  (\bibinfo{year}{2009}) \bibinfo{pages}{289--301}.

\bibitem[{Jiao et~al.(2015)Jiao, Jin, and Lu}]{Jiao2014}
\bibinfo{author}{Y.~Jiao}, \bibinfo{author}{B.~Jin}, \bibinfo{author}{X.~Lu},
  \bibinfo{title}{A primal dual active set with continuation algorithm for the
  $\ell_0$-regularized optimization problem}, \bibinfo{journal}{Appl. Comput.
  Harmon. Anal.} \bibinfo{volume}{39} (\bibinfo{year}{2015})
  \bibinfo{pages}{400--426}.

\bibitem[{Bian and Chen(2020)}]{Bian2019Sparse}
\bibinfo{author}{W.~Bian}, \bibinfo{author}{X.~Chen}, \bibinfo{title}{A
  smoothing proximal gradient algorithm for nonsmooth convex regression with
  cardinality penalty}, \bibinfo{journal}{SIAM J. Numer. Anal.}
  \bibinfo{volume}{58}~(\bibinfo{number}{1}) (\bibinfo{year}{2020})
  \bibinfo{pages}{858--883}.

\bibitem[{Pan et~al.(2017)Pan, Hu, Su, and Yang}]{Pan2017PAMI}
\bibinfo{author}{J.~Pan}, \bibinfo{author}{Z.~Hu}, \bibinfo{author}{Z.~Su},
  \bibinfo{author}{M.-H. Yang}, \bibinfo{title}{${L}_0$-regularized intensity
  and gradient prior for deblurring text images and beyond},
  \bibinfo{journal}{IEEE Trans. Pattern Anal. Mach. Intell.}
  \bibinfo{volume}{39}~(\bibinfo{number}{2}) (\bibinfo{year}{2017})
  \bibinfo{pages}{342--355}.

\bibitem[{Cai et~al.(2019)Cai, Dan, and Zhang}]{Cai2019}
\bibinfo{author}{J.~Cai}, \bibinfo{author}{W.~Dan}, \bibinfo{author}{X.~Zhang},
  \bibinfo{title}{${L}_0$-based sparse canonical correlation analysis with
  application to cross-language document retrieval},
  \bibinfo{journal}{Neurocomputing} \bibinfo{volume}{329}
  (\bibinfo{year}{2019}) \bibinfo{pages}{32--45}.

\bibitem[{Xiong et~al.(2019)Xiong, Zhou, and Qian}]{Xiong2019}
\bibinfo{author}{F.~Xiong}, \bibinfo{author}{J.~Zhou},
  \bibinfo{author}{Y.~Qian}, \bibinfo{title}{Hyperspectral restoration via
  ${L}_0$ gradient regularized low-rank tensor factorization},
  \bibinfo{journal}{IEEE Trans. Geosci. Remote Sens}
  \bibinfo{volume}{57}~(\bibinfo{number}{12}) (\bibinfo{year}{2019})
  \bibinfo{pages}{10410--10425}.

\bibitem[{Osher et~al.(2016)Osher, Ruan, Xiong, Yao, and Yin}]{Osher2016}
\bibinfo{author}{S.~Osher}, \bibinfo{author}{F.~Ruan},
  \bibinfo{author}{J.~Xiong}, \bibinfo{author}{Y.~Yao},
  \bibinfo{author}{W.~Yin}, \bibinfo{title}{Sparse recovery via differential
  inclusions}, \bibinfo{journal}{Appl. Comput. Harmon. Anal.}
  \bibinfo{volume}{41}~(\bibinfo{number}{2}) (\bibinfo{year}{2016})
  \bibinfo{pages}{436--469}.

\bibitem[{Attouch et~al.(2018)Attouch, Chbani, Peypouquet, and
  Redont}]{Attouch2018MP}
\bibinfo{author}{H.~Attouch}, \bibinfo{author}{Z.~Chbani},
  \bibinfo{author}{J.~Peypouquet}, \bibinfo{author}{P.~Redont},
  \bibinfo{title}{Fast convergence of inertial dynamics and algorithms with
  asymptotic vanishing viscosity}, \bibinfo{journal}{Math. Program.}
  \bibinfo{volume}{168}~(\bibinfo{number}{1-2}) (\bibinfo{year}{2018})
  \bibinfo{pages}{123--175}.

\bibitem[{Su et~al.(2016)Su, Boyd, and Candes}]{Su2016}
\bibinfo{author}{W.~Su}, \bibinfo{author}{S.~Boyd}, \bibinfo{author}{E.~J.
  Candes}, \bibinfo{title}{A Differential Equation for Modeling Nesterov's
  Accelerated Gradient Method: Theory and Insights}, \bibinfo{journal}{J. Mach.
  Learn. Res.} \bibinfo{volume}{17}~(\bibinfo{number}{153})
  (\bibinfo{year}{2016}) \bibinfo{pages}{1--43}.

\bibitem[{Attouch and Peypouquet(2016)}]{Attouch2016}
\bibinfo{author}{H.~Attouch}, \bibinfo{author}{J.~Peypouquet},
  \bibinfo{title}{The rate of convergence of Nesterov's accelerated
  forward-backward method is actually faster than $1/k^2$},
  \bibinfo{journal}{SIAM J. Optim.} \bibinfo{volume}{26}~(\bibinfo{number}{3})
  (\bibinfo{year}{2016}) \bibinfo{pages}{1824--1834}.

\bibitem[{Attouch et~al.(2020)Attouch, Chbani, and Riahi}]{Attouch2020}
\bibinfo{author}{H.~Attouch}, \bibinfo{author}{Z.~Chbani},
  \bibinfo{author}{H.~Riahi}, \bibinfo{title}{Convergence rate of inertial
  proximal algorithms with general extrapolation and proximal coefficients},
  \bibinfo{journal}{Vietnam J. Math.} \bibinfo{volume}{48}
  (\bibinfo{year}{2020}) \bibinfo{pages}{247--276}.

\bibitem[{Xia et~al.(2008)Xia, Feng, and Wang}]{Xia2008A}
\bibinfo{author}{Y.~Xia}, \bibinfo{author}{G.~Feng}, \bibinfo{author}{J.~Wang},
  \bibinfo{title}{A novel recurrent Neural Network for solving nonlinear
  optimization problems with inequality constraints}, \bibinfo{journal}{IEEE
  Trans. Neural Netw.} \bibinfo{volume}{19}~(\bibinfo{number}{8})
  (\bibinfo{year}{2008}) \bibinfo{pages}{1340--1353}.

\bibitem[{Gao and Liao(2009)}]{Gao2009A}
\bibinfo{author}{X.~B. Gao}, \bibinfo{author}{L.~Z. Liao}, \bibinfo{title}{A
  new projection-based neural network for constrained variational
  inequalities}, \bibinfo{journal}{IEEE Trans. Neural Netw.}
  \bibinfo{volume}{20}~(\bibinfo{number}{3}) (\bibinfo{year}{2009})
  \bibinfo{pages}{373--388}.

\bibitem[{Hopfield and Tank(1985)}]{Hopfield1985}
\bibinfo{author}{J.~J. Hopfield}, \bibinfo{author}{D.~W. Tank},
  \bibinfo{title}{``Neural'' computation of decisions in optimization
  problems}, \bibinfo{journal}{Biol. Cybern.}
  \bibinfo{volume}{52}~(\bibinfo{number}{3}) (\bibinfo{year}{1985})
  \bibinfo{pages}{141--152}.

\bibitem[{Tank and Hopfield(1986)}]{Tank1988Simple}
\bibinfo{author}{D.~W. Tank}, \bibinfo{author}{J.~J. Hopfield},
  \bibinfo{title}{Simple `Neural' optimization networks: {An} {A/D} converter,
  signal decision circuit, and a linear programming circuit},
  \bibinfo{journal}{IEEE Trans. Circuits Syst.}
  \bibinfo{volume}{33}~(\bibinfo{number}{5}) (\bibinfo{year}{1986})
  \bibinfo{pages}{533--541}.

\bibitem[{Kennedy and Chua(1988)}]{Kennedy1988Neural}
\bibinfo{author}{M.~P. Kennedy}, \bibinfo{author}{L.~O. Chua},
  \bibinfo{title}{Neural Networks for nonlinear programming},
  \bibinfo{journal}{IEEE Trans. Circuits Syst.}
  \bibinfo{volume}{35}~(\bibinfo{number}{5}) (\bibinfo{year}{1988})
  \bibinfo{pages}{554--562}.

\bibitem[{Clemente et~al.(2016)Clemente, Mansour, Ayoubi, Serrano, Mecha,
  Ziade, Falou, and Velazco}]{Clemente2016}
\bibinfo{author}{J.~A. Clemente}, \bibinfo{author}{W.~Mansour},
  \bibinfo{author}{R.~Ayoubi}, \bibinfo{author}{F.~Serrano},
  \bibinfo{author}{H.~Mecha}, \bibinfo{author}{H.~Ziade},
  \bibinfo{author}{W.~E. Falou}, \bibinfo{author}{R.~Velazco},
  \bibinfo{title}{Hardware implementation of a fault-tolerant Hopfield Neural
  Network on FPGAs}, \bibinfo{journal}{Neurocomputing} \bibinfo{volume}{171}
  (\bibinfo{year}{2016}) \bibinfo{pages}{1606--1609}.

\bibitem[{Chen et~al.(2020)Chen, Wang, and Duan}]{Chen2020}
\bibinfo{author}{T.~Chen}, \bibinfo{author}{L.~Wang},
  \bibinfo{author}{S.~Duan}, \bibinfo{title}{Implementation of circuit for
  reconfigurable memristive chaotic neural network and its application in
  associative memory}, \bibinfo{journal}{Neurocomputing} \bibinfo{volume}{380}
  (\bibinfo{year}{2020}) \bibinfo{pages}{36--42}.

\bibitem[{Bian and Xue(2013)}]{Bian2013Neural}
\bibinfo{author}{W.~Bian}, \bibinfo{author}{X.~Xue}, \bibinfo{title}{Neural
  Network for solving constrained convex optimization problems with global
  attractivity}, \bibinfo{journal}{IEEE Trans. Circuits Syst. I-Regul. Pap.}
  \bibinfo{volume}{60}~(\bibinfo{number}{3}) (\bibinfo{year}{2013})
  \bibinfo{pages}{710--723}.

\bibitem[{Yan et~al.(2017)Yan, Fan, and Wang}]{Yan2017A}
\bibinfo{author}{Z.~Yan}, \bibinfo{author}{J.~Fan}, \bibinfo{author}{J.~Wang},
  \bibinfo{title}{A collective neurodynamic approach to constrained global
  optimization}, \bibinfo{journal}{IEEE Trans. Neural Netw. Learn. Syst.}
  \bibinfo{volume}{28}~(\bibinfo{number}{5}) (\bibinfo{year}{2017})
  \bibinfo{pages}{1206--1215}.

\bibitem[{Le and Wang(2017)}]{Le2017A}
\bibinfo{author}{X.~Le}, \bibinfo{author}{J.~Wang}, \bibinfo{title}{A
  two-time-scale neurodynamic approach to constrained minimax optimization},
  \bibinfo{journal}{IEEE Trans. Neural Netw. Learn. Syst.}
  \bibinfo{volume}{28}~(\bibinfo{number}{3}) (\bibinfo{year}{2017})
  \bibinfo{pages}{620--629}.

\bibitem[{Bian et~al.(2018)Bian, Ma, Qin, and Xue}]{BianNN2018}
\bibinfo{author}{W.~Bian}, \bibinfo{author}{L.~Ma}, \bibinfo{author}{S.~Qin},
  \bibinfo{author}{X.~Xue}, \bibinfo{title}{Neural Network for nonsmooth
  pseudoconvex optimization with general convex constraints},
  \bibinfo{journal}{Neural Netw.} \bibinfo{volume}{101} (\bibinfo{year}{2018})
  \bibinfo{pages}{1--14}.

\bibitem[{Shen et~al.(2012)Shen, Pan, and Zhu}]{Shen2012Likelihood}
\bibinfo{author}{X.~Shen}, \bibinfo{author}{W.~Pan}, \bibinfo{author}{Y.~Zhu},
  \bibinfo{title}{Likelihood-based selection and sharp parameter estimation},
  \bibinfo{journal}{J. Amer. Statist. Assoc.}
  \bibinfo{volume}{107}~(\bibinfo{number}{497}) (\bibinfo{year}{2012})
  \bibinfo{pages}{223--232}.

\bibitem[{Zheng et~al.(2014)Zheng, Fan, and Lv}]{Zheng2014High}
\bibinfo{author}{Z.~Zheng}, \bibinfo{author}{Y.~Fan}, \bibinfo{author}{J.~Lv},
  \bibinfo{title}{High dimensional thresholded regression and shrinkage
  effect}, \bibinfo{journal}{J. R. Stat. Soc. Ser. B. Sta. Meth.}
  \bibinfo{volume}{76}~(\bibinfo{number}{3}) (\bibinfo{year}{2014})
  \bibinfo{pages}{627--649}.

\bibitem[{Foucart and Lai(2009)}]{Foucart2009Sparsest}
\bibinfo{author}{S.~Foucart}, \bibinfo{author}{M.~J. Lai},
  \bibinfo{title}{Sparsest solutions of underdetermined linear systems via
  $\ell_q$-minimization for $0<q\leq1$}, \bibinfo{journal}{Appl. Comput.
  Harmon. Anal.} \bibinfo{volume}{26}~(\bibinfo{number}{3})
  (\bibinfo{year}{2009}) \bibinfo{pages}{395--407}.

\bibitem[{Zhang(2013)}]{Zhang2013Multi}
\bibinfo{author}{T.~Zhang}, \bibinfo{title}{Multi-stage convex relaxation for
  feature selection}, \bibinfo{journal}{Bernoulli}
  \bibinfo{volume}{19}~(\bibinfo{number}{5B}) (\bibinfo{year}{2013})
  \bibinfo{pages}{2277--2293}.

\bibitem[{Fan and Li(2001)}]{Fan2001Variable}
\bibinfo{author}{J.~Fan}, \bibinfo{author}{R.~Li}, \bibinfo{title}{Variable
  selection via nonconvave penalized likelihood and its oracle properties},
  \bibinfo{journal}{J. Amer. Statist. Assoc.}
  \bibinfo{volume}{96}~(\bibinfo{number}{456}) (\bibinfo{year}{2001})
  \bibinfo{pages}{1348--1360}.

\bibitem[{Zhang(2010)}]{Zhang2010NEARLY}
\bibinfo{author}{C.~Zhang}, \bibinfo{title}{Nearly unbiased variable selection
  under minimax concave penalty}, \bibinfo{journal}{Ann. Stat.}
  \bibinfo{volume}{38}~(\bibinfo{number}{2}) (\bibinfo{year}{2010})
  \bibinfo{pages}{894--942}.

\bibitem[{Xu et~al.(2012)Xu, Chang, Xu, and Zhang}]{Xu2012L1}
\bibinfo{author}{Z.~Xu}, \bibinfo{author}{X.~Chang}, \bibinfo{author}{F.~Xu},
  \bibinfo{author}{H.~Zhang}, \bibinfo{title}{${L}_{1/2}$ regularization: a
  thresholding representation theory and a fast solver}, \bibinfo{journal}{IEEE
  Trans. Neural Netw. Learn. Syst.} \bibinfo{volume}{23}~(\bibinfo{number}{7})
  (\bibinfo{year}{2012}) \bibinfo{pages}{1013--1027}.

\bibitem[{Bian and Chen(2012)}]{Wei2012Smoothing}
\bibinfo{author}{W.~Bian}, \bibinfo{author}{X.~Chen}, \bibinfo{title}{Smoothing
  Neural Network for constrained non-{Lipschitz} optimization with
  applications}, \bibinfo{journal}{IEEE Trans. Neural Netw. Learn. Syst.}
  \bibinfo{volume}{23}~(\bibinfo{number}{3}) (\bibinfo{year}{2012})
  \bibinfo{pages}{399--411}.

\bibitem[{Bian and Chen(2014)}]{Bian2014Neural}
\bibinfo{author}{W.~Bian}, \bibinfo{author}{X.~Chen}, \bibinfo{title}{Neural
  Network for nonsmooth, nonconvex constrained minimization via smooth
  approximation}, \bibinfo{journal}{IEEE Trans. Neural Netw. Learn. Syst.}
  \bibinfo{volume}{25}~(\bibinfo{number}{3}) (\bibinfo{year}{2014})
  \bibinfo{pages}{545--556}.

\bibitem[{Li et~al.(2020)Li, Bian, and Xue}]{Li2019}
\bibinfo{author}{W.~Li}, \bibinfo{author}{W.~Bian}, \bibinfo{author}{X.~Xue},
  \bibinfo{title}{Projected neural network for a class of non-Lipschitz
  optimization problems with linear constraints}, \bibinfo{journal}{IEEE Trans.
  Neural Netw. Learn. Syst.} \bibinfo{volume}{31}~(\bibinfo{number}{9})
  (\bibinfo{year}{2020}) \bibinfo{pages}{3361--3373}.

\bibitem[{Soubies et~al.(2017)Soubies, Blanc-F\'eraud, and
  Aubert}]{Soubies2017A}
\bibinfo{author}{E.~Soubies}, \bibinfo{author}{L.~Blanc-F\'eraud},
  \bibinfo{author}{G.~Aubert}, \bibinfo{title}{A unified view of exact
  continuous penalties for $\ell_2$-$\ell_0$ minimization},
  \bibinfo{journal}{SIAM J. Optim.} \bibinfo{volume}{27}~(\bibinfo{number}{3})
  (\bibinfo{year}{2017}) \bibinfo{pages}{2034--2060}.

\bibitem[{Fung and Mangasarian(2011)}]{2011Equivalence}
\bibinfo{author}{G.~M. Fung}, \bibinfo{author}{O.~L. Mangasarian},
  \bibinfo{title}{Equivalence of minimal $\ell_0-$ and $\ell_p-$ norm solutions
  of linear equalities, inequalities and linear programs for sufficiently small
  $p$}, \bibinfo{journal}{J. Optim. Theory Appl.}
  \bibinfo{volume}{151}~(\bibinfo{number}{1}) (\bibinfo{year}{2011})
  \bibinfo{pages}{1--10}.

\bibitem[{Zhao et~al.(2020)Zhao, He, Huang, Huang, and Li}]{Zhao2020}
\bibinfo{author}{Y.~Zhao}, \bibinfo{author}{X.~He}, \bibinfo{author}{T.~Huang},
  \bibinfo{author}{J.~Huang}, \bibinfo{author}{P.~Li}, \bibinfo{title}{A
  smoothing neural network for minimization $l_1$-$l_p$ in sparse signal
  reconstruction with measurement noises}, \bibinfo{journal}{Neural Netw.}
  \bibinfo{volume}{122} (\bibinfo{year}{2020}) \bibinfo{pages}{40--53}.

\bibitem[{Feng et~al.(2017)Feng, Leung, Constantinides, and Zeng}]{Feng2017}
\bibinfo{author}{R.~Feng}, \bibinfo{author}{C.-S. Leung},
  \bibinfo{author}{A.~G. Constantinides}, \bibinfo{author}{W.-J. Zeng},
  \bibinfo{title}{Lagrange programming neural network for nondifferentiable
  optimization problems in sparse approximation}, \bibinfo{journal}{IEEE Trans.
  Neural Netw. Learn. Syst.} \bibinfo{volume}{28}~(\bibinfo{number}{10})
  (\bibinfo{year}{2017}) \bibinfo{pages}{2395--2407}.

\bibitem[{Coddington and Levinson(1955)}]{Coddington1984Theory}
\bibinfo{author}{E.~A. Coddington}, \bibinfo{author}{N.~Levinson},
  \bibinfo{title}{Theory of Ordinary Differential Equations},
  \bibinfo{publisher}{New York: McGraw-Hill Book Co., Inc.},
  \bibinfo{year}{1955}.

\bibitem[{Kinderlehrer and Stampacchia(1980)}]{Kinderlehrer1980An}
\bibinfo{author}{D.~Kinderlehrer}, \bibinfo{author}{G.~Stampacchia},
  \bibinfo{title}{An Introduction to Variational Inequalities and Their
  Applications}, \bibinfo{publisher}{New York: Academic}, \bibinfo{year}{1980}.

\bibitem[{Clarke(1983)}]{Clarke1983Optimization}
\bibinfo{author}{F.~H. Clarke}, \bibinfo{title}{Optimization and Nonsmooth
  Analysis}, \bibinfo{publisher}{New York: Wiley}, \bibinfo{year}{1983}.

\bibitem[{Hale(1980)}]{Hale1980}
\bibinfo{author}{J.~K. Hale}, \bibinfo{title}{Ordinary Differential Equations},
  \bibinfo{publisher}{New York: Wiley}, \bibinfo{year}{1980}.

\bibitem[{Aubin and Cellina(1984)}]{Aubin1984Differential}
\bibinfo{author}{J.~P. Aubin}, \bibinfo{author}{A.~Cellina},
  \bibinfo{title}{Differential Inclusions: Set-Valued Maps and Viability
  Theory}, \bibinfo{publisher}{New York: Springer-Verlag},
  \bibinfo{year}{1984}.

\bibitem[{Chen et~al.(2012)Chen, Ng, and Zhang}]{Chen2012Non}
\bibinfo{author}{X.~Chen}, \bibinfo{author}{M.~K. Ng},
  \bibinfo{author}{C.~Zhang}, \bibinfo{title}{Non-{Lipschitz}
  $\ell_p$-regularization and box constrained model for image restoration},
  \bibinfo{journal}{IEEE Trans. Image Process.}
  \bibinfo{volume}{21}~(\bibinfo{number}{12}) (\bibinfo{year}{2012})
  \bibinfo{pages}{4709--4721}.

\bibitem[{Bian and Chen(2015)}]{Bian2015Optimality}
\bibinfo{author}{W.~Bian}, \bibinfo{author}{X.~Chen},
  \bibinfo{title}{Optimality and complexity for constrained optimization
  problems with nonconvex regularization}, \bibinfo{journal}{Math. Oper. Res.}
  \bibinfo{volume}{42}~(\bibinfo{number}{4}) (\bibinfo{year}{2015})
  \bibinfo{pages}{1063--1084}.

\bibitem[{Chartrand and Staneva(2008)}]{Chartrand2008Restricted}
\bibinfo{author}{R.~Chartrand}, \bibinfo{author}{V.~Staneva},
  \bibinfo{title}{Restricted isometry properties and nonconvex compressive
  sensing}, \bibinfo{journal}{Inverse Probl.}
  \bibinfo{volume}{24}~(\bibinfo{number}{3}) (\bibinfo{year}{2008})
  \bibinfo{pages}{657--682}.

\bibitem[{Huang et~al.(2008)Huang, Horowitz, and Ma}]{Huang2008Asymptotic}
\bibinfo{author}{J.~Huang}, \bibinfo{author}{J.~L. Horowitz},
  \bibinfo{author}{S.~Ma}, \bibinfo{title}{Asymptotic properties of bridge
  estimators in sparse high-dimensional regression models},
  \bibinfo{journal}{Ann. Stat.} \bibinfo{volume}{36}~(\bibinfo{number}{2})
  (\bibinfo{year}{2008}) \bibinfo{pages}{587--613}.

\bibitem[{Li et~al.(2010)Li, Song, and Wu}]{Li2010Generalized}
\bibinfo{author}{G.~Li}, \bibinfo{author}{S.~Song}, \bibinfo{author}{C.~Wu},
  \bibinfo{title}{Generalized gradient projection neural networks for nonsmooth
  optimization problems}, \bibinfo{journal}{Sci. Chin. Inf. Sci.}
  \bibinfo{volume}{53}~(\bibinfo{number}{5}) (\bibinfo{year}{2010})
  \bibinfo{pages}{990--1005}.

\bibitem[{Chen(2012)}]{Chen-MP}
\bibinfo{author}{X.~Chen}, \bibinfo{title}{Smoothing methods for nonsmooth,
  nonconvex minimization}, \bibinfo{journal}{Math. Program.}
  \bibinfo{volume}{134}~(\bibinfo{number}{1}) (\bibinfo{year}{2012})
  \bibinfo{pages}{71--99}.

\bibitem[{Stamey et~al.(1989)Stamey, Kabalin, McNeal, Johnstone, Freiha,
  Redwine, and Yang}]{Stamey1989Prostate}
\bibinfo{author}{T.~Stamey}, \bibinfo{author}{J.~Kabalin},
  \bibinfo{author}{J.~McNeal}, \bibinfo{author}{I.~Johnstone},
  \bibinfo{author}{F.~Freiha}, \bibinfo{author}{E.~Redwine},
  \bibinfo{author}{N.~Yang}, \bibinfo{title}{Prostate specific antigen in the
  diagnosis and treatment of adenocarcinoma of the prostate. {II}. Radical
  prostatectomy treated patients}, \bibinfo{journal}{J. Urology}
  \bibinfo{volume}{141} (\bibinfo{year}{1989}) \bibinfo{pages}{1076--1083}.

\end{thebibliography}


\end{document}